\documentclass[english]{IEEEtran}

\usepackage[T1]{fontenc}
\usepackage[latin9]{inputenc}
\usepackage{color}
\usepackage[dvipsnames]{xcolor}
\usepackage{float}
\usepackage{textcomp}
\usepackage{mathrsfs}
\usepackage{mathtools}
\usepackage{amsmath}
\usepackage{amsthm}
\usepackage{amssymb}
\usepackage{stmaryrd}
\usepackage{stackrel}
\usepackage{graphicx}
\usepackage{wasysym}
\usepackage{url}
\makeatletter

\floatstyle{ruled}
\newfloat{algorithm}{tbp}{loa}
\providecommand{\algorithmname}{Algorithm}
\floatname{algorithm}{\protect\algorithmname}

\theoremstyle{plain}
\newtheorem{thm}{\protect\theoremname}
\theoremstyle{definition}
\newtheorem{defn}{\protect\definitionname}
\theoremstyle{definition}
\newtheorem{problem}{\protect\problemname}
\theoremstyle{plain}
\newtheorem{lem}{\protect\lemmaname}
\theoremstyle{definition}
\newtheorem{example}{\protect\examplename}
\theoremstyle{remark}

\theoremstyle{plain}

\providecommand{\lemmaname}{Lemma}
\usepackage{cite}
\usepackage{algpseudocode}
\usepackage{setspace}
\IEEEoverridecommandlockouts
\makeatother

\usepackage{babel}
\providecommand{\definitionname}{Definition}
\providecommand{\examplename}{Example}
\providecommand{\problemname}{Problem}
\providecommand{\theoremname}{Theorem}

\usepackage{color}

\begin{document}
\title{Online Motion Planning with Soft Metric Interval Temporal Logic in Unknown Dynamic Environment \thanks{Z. Li, and Z. Kan are with the Department of Automation, University
of Science and Technology of China, Hefei, China. M. Cai is with the
Department of Mechanical Engineering and Mechanics, Lehigh University,
Bethlehem, PA, USA. SP. Xiao is with the Department of Mechanical Engineering, University of Iowa, Iowa City, USA.}}
\author{Zhiliang Li, Mingyu Cai, Shaoping Xiao, Zhen Kan}
\maketitle
\begin{abstract}
Motion planning of an autonomous system with high-level specifications has wide applications. However, research of formal languages involving timed temporal logic is still under investigation.
Furthermore, many
existing results rely on a key assumption that user-specified tasks
are feasible in the given environment.
Challenges arise when the
operating environment is dynamic and unknown since the environment
can be found prohibitive, leading to potentially conflicting tasks where pre-specified timed missions cannot be fully satisfied. Such issues become even more challenging when considering time-bound requirements.
To
address these challenges, this work proposes a control framework
that considers hard constraints to enforce safety
requirements and soft constraints to enable task relaxation. The metric interval temporal logic (MITL) specifications are
employed to deal
with time-bound constraints. By constructing a relaxed timed product automaton, an online
motion planning strategy is synthesized with a receding horizon controller
to generate policies, achieving multiple objectives in decreasing order of priority 1) formally guarantee the satisfaction of hard safety constraints; 2)
mostly fulfill soft timed tasks; and 3) collect time-varying rewards
as much as possible. Another novelty of the relaxed structure is to consider violations of both time and tasks for infeasible cases.  
Simulation results are provided
to validate the proposed approach.

\global\long\def\Dist{\operatorname{Dist}}%
\global\long\def\Inf{\operatorname{Inf}}%
\global\long\def\Sense{\operatorname{Sense}}%
\global\long\def\Eval{\operatorname{Eval}}%
\global\long\def\Info{\operatorname{Info}}%
\global\long\def\ResetRabin{\operatorname{ResetRabin}}%
\global\long\def\Post{\operatorname{Post}}%
\global\long\def\Acc{\operatorname{Acc}}%
\global\long\def\Type{\operatorname{Type}}%
\end{abstract}


\section{INTRODUCTION}

Complex rules in modern tasks often specify desired
system behaviors and timed temporal constraints that require mission
completion within a given period. Performing such tasks can be challenging,
especially when the operating environment is dynamic and unknown.
For instance, user-specified missions or temporal constraints can
be found infeasible during motion planning. Therefore,
this work is motivated for online motion planning subject to timed high-level specifications.

Linear temporal logic (LTL) has been widely used
for task and motion planning due to its rich expressivity and resemblance
to natural language \cite{Belta2007}. When considering
timed formal language, as an extension of traditional LTL, timed temporal languages
such as metric interval temporal logic (MITL) \cite{Alur1996}, signal temporal logic (STL) \cite{Maler2004}, time-window temporal logic (TWTL) \cite{Vasile2017TWTL}, are often employed. However, most existing results are
built on the assumption that user-specified tasks are feasible. New
challenges arise when the operating environment is dynamic and unknown
since the environment can become prohibitive (e.g., an area to be
visited is found later to be surrounded by obstacles), leading to
mission failure. 

To address these challenges, tasks with temporal
logic specifications are often relaxed to be fulfilled
as much as possible.
A least-violating control strategy is developed in \cite{Castro2013,Tumova2016,lahijanian2016iterative, Vasile2017,Cai2020b,Cai2021_soft_RL}
 to enforce the revised
motion planning close to the original LTL specifications. In \cite{Guo2015,Andersson2018,Ahlberg2019},
hard and soft constraints are considered so that the satisfaction of
hard constraints is guaranteed while soft constraints are minimally
violated. Time relaxation of TWTL has been investigated in \cite{Peterson2020,kamale2021automata,aksaray2021learning}.
Receding horizon control (RHC) is also integrated with
temporal logic specifications to deal with motion planning in dynamic
environments \cite{wongpiromsarn2012receding,Ding2014,ulusoy2014receding, Lu2018,Cai2020c,Aasi2021}. Other
representative results include learning-based methods \cite{Hasanbeig2019reinforcement, Cai2020, Cai2020d, Cai2021modular, Cai2021safe}
and sampling-based reactive methods \cite{Vasile2020,kantaros2020reactive}.
Most of the results mentioned above do not consider time constraints
in motion planning. 
MITL is an automaton-based temporal logic that has flexibility to express general time constraints. Recent works \cite{Nikou2016,Nikou2018, verginis2019, li2021policy, Xu2021controller}  propose different strategies to satisfy MITL formulas. The works of \cite{Nikou2016,Nikou2018} consider cooperative planning of a multi-agent system with MITL specifications and the work of \cite{Xu2021controller} further investigates MITL planning of a MAS subject to intermittent communication. When considering dynamic environments, MITL with probabilistic distributions is developed in \cite{li2021policy} to express time-sensitive missions, and a Reconfigurable algorithm is developed in \cite{verginis2019}. However, the aforementioned works assume that the desired MTIL specifications are always feasible for the robotic system. Sofie et al. \cite{Andersson2018,Ahlberg2019} first take into account the soft MITL constraints and studies the interactions of human-robot,
but only static environments are considered. It is not yet understood how timed temporal tasks
can be successfully managed in a dynamic and unknown environment,
where predefined tasks may be infeasible.

% Although the works \cite{Andersson2018,Ahlberg2019}
% take into account time constraints, only purely static environments
% are considered. It is not yet understood how timed temporal tasks
% can be successfully managed in a dynamic and unknown environment,
% where predefined tasks may be infeasible.

Motivated by these challenges, this work considers
online motion planning of an autonomous system with timed temporal
specifications. Unlike STL defined over predicates, MITL provides more general time constraints and can express tasks over infinite horizons. Furthermore, MITL can be translated into timed automata that allow us to exploit graph-theoretical approaches for analysis and design. Therefore, MITL is used in this work. 

The contributions of this work are multi-fold. First,
the operating environment is not fully known a
priori and dynamic in the sense of containing
mobile obstacles and time-varying areas of interest that can only
be observed locally. The dynamic and unknown environment
can lead to potentially conflicting tasks (i.e., the pre-specified
MITL missions or time constraints cannot be fully
satisfied). Inspired by our previous work \cite{Cai2020c},
we consider both hard and soft constraints.
The motivation behind this design is that safety is crucial in real-world applications; therefore, we formulate safety requirements (e.g., avoid obstacles) as hard constraints that cannot be violated in all cases. In contrast, soft constraints can be relaxed if the environment does not permit such specifications so that the agent can accomplish the tasks as much as possible.
% Hard constraints enforce safety requirements
% (e.g., avoid obstacles), while soft constraints represent tasks that
% can be relaxed not strictly to follow the specifications if the environment
% does not permit. 
Second, to deal with time constraints, we apply MITL specifications to
model timed temporal tasks and further classify soft constraints by
how they can be violated. For instance, the mission can fail because the agent cannot reach the destination on time, or 
the agent visits some risky regions. Therefore,
the innovation considers violations of both time constraints and task specifications caused by dynamic obstacles, which can be formulated as continuous and discrete types, respectively.

Our framework is to generate controllers achieving multiple objectives
in decreasing order of priority: 1) formally guarantee the satisfaction
of hard constraints; 2) mostly satisfy soft constraints (i.e., minimizing
the violation cost); and 3) collect time-varying rewards as much as
possible (e.g., visiting areas of higher interest more often). 
% We construct a novel relaxed timed product
% automaton to alleviate these challenges, allowing the agent not to follow the desired
% MITL constraints strictly. An online motion planning strategy is synthesized
% with a receding horizon controller to adapt to the dynamic environment.
% By solving an optimization problem online, the maximum satisfaction of the
% specification is guaranteed, and an optimal accumulated reward is obtained. It's worth noting that the RHC only considers local dynamic information online while global satisfaction is formally guaranteed, which is efficient for large-scale environments.
% Simulation results are provided to validate the proposed approach. 
Different from \cite{Ding2014} that assumes the LTL specifications can be exactly achieved, we relax the assumption and consider tasks with time constraints described by MITL formulas. Unlike \cite{Andersson2018,Ahlberg2019}, we consider a dynamic unknown environment where the agent needs to detect and update in real-time. In particular, a multi-objective RHC is synthesized online to adapt to the dynamic environment, which guarantees the safety constraint and minimum violation of the soft specification.
 Furthermore, it's worth noting that the RHC only considers local dynamic information online while global satisfaction is formally guaranteed, which is efficient for large-scale environments. Finally, we demonstrate the effectiveness of our algorithm by a complex infinite task in simulation.

\section{PRELIMINARIES\label{Sec:Preliminary}}

A dynamical system with finite states evolving in an environment can
be modeled by a weighted transition system.
\begin{defn}
\cite{Baier2008}\label{def:WTS} A weighted transition system (WTS)
is a tuple $\mathscr{\mathcal{T=\textrm{\ensuremath{\left(Q,q_{0},\delta,\mathcal{AP},L,\mathcal{\omega}\right)}}}}$,
where $Q$ is a finite set of states; $q_{0}\in Q$ is the initial
state; $\delta\in Q\times Q$ is the state transitions; $\mathcal{AP}$
is the finite set of atomic propositions; $L:Q\rightarrow2^{\mathcal{AP}}$
is a labeling function, and $\omega:\delta\rightarrow\mathbb{R}^{+}$
assigns a positive weight to each transition.
\end{defn}
A timed run of a WTS $\mathcal{T}$ is an infinite sequence $\boldsymbol{r}=(q_{0},\tau_{0})(q_{1},\tau_{1})\ldots$,
where $\boldsymbol{q}=q_{0}q_{1}\ldots$ is a trajectory with $q_{i}\in Q$
, and $\boldsymbol{\tau}=\tau_{0}\tau_{1}\ldots$ is a time sequence
with $\tau_{0}=0$ and $\tau_{i+1}=\tau_{i}+\omega(q_{i},q_{i+1}),\forall i\geq0$.
The timed run $\boldsymbol{r}$ generates a timed word $\boldsymbol{w}=(\sigma_{0},\tau_{0})(\sigma_{1},\tau_{1})\ldots$
where $\boldsymbol{\sigma}=\sigma_{0}\sigma_{1}\ldots$ is an infinite
word with $\sigma_{i}=L(q_{i})$ for $i\geq0$. Let
$R_{k}(q)$ denote the time-varying reward associated with a state
$q$ at time $k$. The reward reflects the time-varying objective in the environment.
Given a predicted trajectory   $\boldsymbol{q}_{k}=q_{0}q_{1}\ldots q_{N}$
at time $k$ with a finite horizon $N$, the accumulated reward along the
trajectory $\boldsymbol{q}_{k}$ can be computed as  $\boldsymbol{R}_{k}(\boldsymbol{q}_{k})=\sum_{i=1}^{N}R_{k}(q_{i})$. 

Note that this paper mainly studies high-level planning and decision-making problems. Similar to \cite{Cai2020b}, we assume low-level controllers can achieve go-to-goal navigation, which can be abstracted by WTS. We further assume that the workspace boundaries are known, which is a common assumption in many existing works \cite{Nikou2016,Nikou2018,Guo2015,Andersson2018,Ahlberg2019,Ding2014}.

\subsection{Metric Interval Temporal Logic}\label{sec:MITL}

Metric interval temporal logic (MITL) is a specific temporal logic
that includes timed temporal specification \cite{Nikou2018}. The
syntax of MITL formulas is defined as $\phi:=p\mid\neg\phi\mid\phi_{1}\land\phi_{2}\mid\diamondsuit_{I}\phi\mid\boxempty_{I}\phi\mid\phi_{1}\mathcal{U_{\mathit{I}}}\phi_{2}$,
where $p\in\mathcal{AP}$, $\land(\textrm{conjunction}),\lnot(\textrm{negation})$
are Boolean operators and $\diamondsuit_{I}\text{(eventually)}$,
$\boxempty_{I}(\textrm{always})$ , $\mathcal{U_{\mathit{I}}}(\textrm{until})$
are temporal operators bounded by the non-empty time interval $I=[a,b]$ with $a,b\in\mathbb{R}_{\geq0},b>a$. They are called temporally bounded operators if $b\neq\infty$, and non-temporally
bounded operators otherwise. A formula $\phi$ containing a temporally
bounded operator will be called a temporally bounded formula. The same holds
for non-temporally bounded formulas.  

Given a timed run $\boldsymbol{r}$ of $\mathcal{T}$ and an MITL formula $\phi$, let 
$(\boldsymbol{r},i)$ denote the indexed element $(q_{i},\tau_{i})$. Then the satisfaction relationship $\models$ of MITL can be defined as:

\[
\begin{array}{l}
(\boldsymbol{r},i)\models p\Longleftrightarrow p\in L(q_{i})\\
(\boldsymbol{r},i)\models\lnot\phi\Longleftrightarrow(\boldsymbol{r},i)\nvDash\phi\\
(\boldsymbol{r},i)\models\phi_{1}\land\phi_{2}\Longleftrightarrow(\boldsymbol{r},i)\models\phi_{1}\textrm{\textrm{ and }}(\boldsymbol{r},i)\models\phi_{2}\\
(\boldsymbol{r},i)\models\diamondsuit_{I}\phi\Longleftrightarrow\exists j,i\leq j,s.t.(\boldsymbol{r},j)\models\phi,\tau_{j}-\tau_{i}\in I\\
(\boldsymbol{r},i)\models\boxempty_{I}\phi\Longleftrightarrow\forall j,i\leq j,\tau_{j}-\tau_{i}\in I\Rightarrow(\boldsymbol{r},j)\models\phi\\
(\boldsymbol{r},i)\models\phi_{1}\mathscr{\mathcal{U}}_{I}\phi_{2}\Longleftrightarrow\exists j,i\leq j,s.t.(\boldsymbol{r},j)\models\phi_{2},\tau_{j}-\tau_{i}\\
\textrm{\ \ \ \ \ \ \ \ \ \ \ \  \ensuremath{\in I} and }(\boldsymbol{r},k)\models\phi_{1}\textrm{\textrm{ for }\textrm{every} }i\leq k\leq j
\end{array}
\]

\subsection{Timed B\"uchi Automaton\label{subsec:TBA}} 

Let $X=\{x_{1},x_{2},\ldots,x_{M}\}$ be a finite set of clocks. The
set of clock constraints $\Phi(X)$ is defined by the grammar $\varphi\coloneqq\top\mid\lnot\varphi\mid\varphi_{1}\wedge\varphi_{2}\mid x\Join c$,
where $x\in X$ is a clock, $c\in\mathbb{R^{+}}$ is a clock constant
and $\Join\in\{<,>,\ge,\le,=\}$. A clock valuation $\nu$ : $X\rightarrow\mathbb{R^{+}}$
assigns a real value to each clock. We denote by $\nu\models\varphi$
if the valuation $\nu$ satisfies the clock constraint $\varphi$,
where $\nu=(\nu_{1},\ldots,\nu_{M})$ with $\nu_{i}$ being the valuation
of $x_{i}$, $\forall i\in{1,\ldots,M}$. An MITL formula can be converted into a Timed B\"uchi Automaton (TBA) \cite{Alur1994}. 

\begin{defn}
A TBA is a tuple $\mathcal{A}=(S,S_{0},\mathcal{AP},\mathcal{L},X,I_{X},E,F)$
where $S$ is a finite set of states; $S_{0}\subseteq S$ is the set
of initial states; $2^{\mathcal{AP}}$ is the alphabet where $\mathcal{AP}$
is a finite set of atomic propositions; $\mathcal{L}:S\rightarrow2^{\mathcal{AP}}$ is a labeling function; $X$ is a finite set of clocks;
% $I_{X}:S\rightarrow\Phi_{X}$ \textcolor{red}{(Xiao: $\phi(X)$?)} is a map from states to clock constraints;
$I_{X}:S\rightarrow\Phi(X)$ is a map from states to clock constraints;
% $E\subseteq S\times\Phi(X)\times2^{X}\times S$
$E\subseteq S\times\Phi(X)\times2^{\mathcal{AP}}\times S$
 represents the set
of edges of form $e=(s,g,a,s^{\prime})$ where $s,s^{\prime}$
are the source and target states, $g$ is the guard of edge via an assigned clock constraint, and $a\in2^{\mathcal{AP}}$
is an input symbol; $F\subseteq S$ is a set of accepting states.
\end{defn}

\begin{defn}
An automata timed run $\boldsymbol{r}_{\mathcal{A}}=(s_{0},\tau_{0})\ldots(s_{n},\tau_{n})$
of a TBA $\mathcal{A}$, corresponding to the timed run $\boldsymbol{r}=(q_{0},\tau_{0})\ldots(q_{n},\tau_{n})$ of a WTS $\mathcal{T}$,
is a sequence where $s_{0}\in S_{0}$, $s_{j}\in S$, and $(s_{j},g_{j},a_{j},s_{j+1})\in E\ \forall j\geq0$
such that i) $\tau_{j}\models g_{j},j\geq0,$ and ii) $L(q_{j})\subseteq\mathcal{L}(s_{j}),\forall j$.
\end{defn}

\begin{defn}
Given a WTS $\mathcal{T}=\textrm{\ensuremath{\left(Q,q_{0},\delta,\mathcal{AP},L,\mathcal{\omega}\right)}}$
and a TBA $\mathcal{A}=(S,S_{0},\mathcal{AP},\mathcal{L},X,I_{X},E,F)$, the product
automaton $\mathcal{P}=\mathcal{T\times A}$ is defined as a tuple
$\mathcal{P}=\{P,P_{0},\mathcal{AP},L_{\mathcal{P}},\delta_{\mathcal{\mathcal{P}}},I_{X}^{\mathcal{P}}\mathcal{\text{,}F_{\mathcal{P}}},\omega_{\mathcal{P}}\}$,
where $P\subseteq\left\{ (q,s)\in Q\times S:L(q)\subseteq \mathcal{L}(s)\right\}$ is the set of states; $P_{0}=\{q_{0}\}\times S_{0}$ is the
set of initial states; $L_{\mathcal{P}}=P\rightarrow2^{\mathcal{AP}}$
is a labeling function, i.e., $L_{\mathcal{P}}(p)=L(q)$;
$\delta_{\mathcal{P}}\subseteq P\times P$ 
is the set of transitions defined such that $((q,s),(q^{\prime},s^{\prime}))\in\delta_{\mathcal{P}}$
if and only if ($q,q^{\prime})\in\delta$ and $\exists g,a,$
such that $(s,g,a,s^{\prime})\in E$; $I_{X}^{\mathcal{P}}(p)=I_{X}(s)$
is a map of clock constraints; $\mathcal{F}_{\mathcal{P}}=Q\times F$
is the set of accepting states; $\omega_{\mathcal{P}}\colon\delta_{\mathcal{P}}\rightarrow\mathbb{R}^{+}$
is the positive weight function, i.e., $\omega_{\mathcal{P}}(p,p^{\prime})=\omega(q,q^{\prime})$.
\end{defn}

\section{Problem Formulation\label{sec:PF}}

To better explain our motion planning strategy, we use the following running
example throughout this work.

\begin{example}
\label{examp1}
\begin{figure}
\centering{}\includegraphics[scale=0.37]{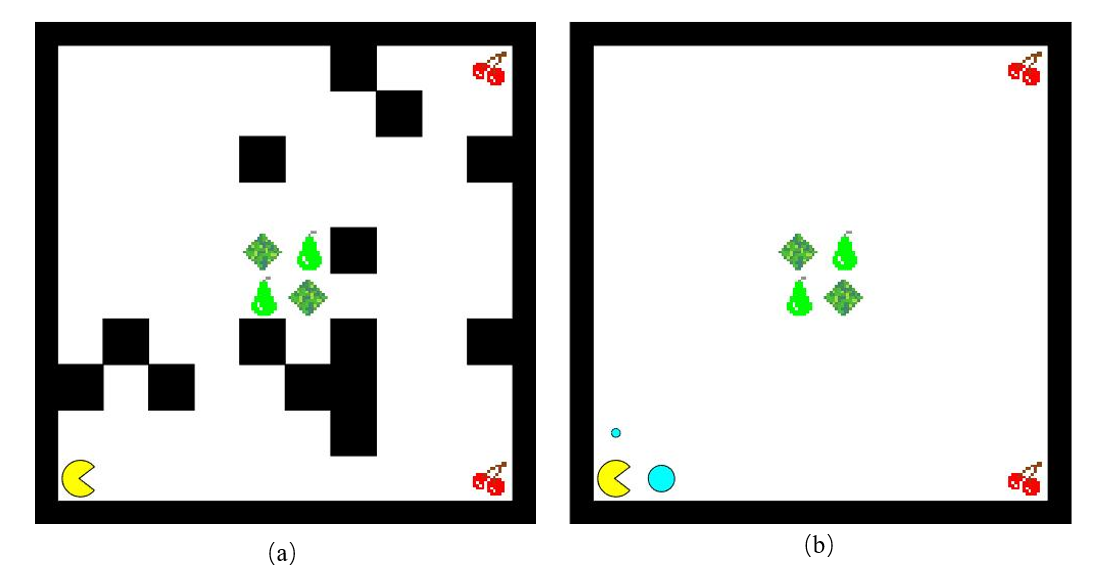}\caption{\label{fig:example}(a) The simplified Pac-Man game with randomly
populated $\mathtt{pear}$, $\mathtt{cherry}$, $\mathtt{grass}$
and $\mathtt{obstacle}$ (i.e., black blocks). (b) The sensed environment
by Pac-Man initially. Pac-man only knows the positions of $\mathtt{pear}$,
$\mathtt{cherry}$, $\mathtt{grass}$ and locally sensed time-varying
rewards (i.e., cyan dots), without any knowledge about the number
and distribution of obstacles.}
\end{figure}
 Consider a motion planning  problem for a simplified Pac-Man game
in Fig. \ref{fig:example}. The maze is abstracted to a named grid-like
graph, and the set of atomic propositions $\mathcal{AP}=\left\{ \mathtt{obstacle,grass,pear,cherry}\right\} $
indicates the labeled properties of regions. In particular, $\mathtt{obstacle}$
represents areas that should be totally avoided, $\mathtt{grass}$
represents risky areas that should be avoided if possible, and $\mathtt{pear}$
and $\mathtt{cherry}$ represent points of interest. The environment
is dynamic in the sense of containing mobile obstacles and time-varying
rewards $R_{k}(q)\in\mathbb{R}^{+}$ that are randomly generated.
Cyan dots represent the rewards with size proportional to
their value. 

We make the following assumptions: 1) the environment is
only partially known to Pac-Man, i.e., the locations of $\mathtt{pear}$,
$\mathtt{cherry}$, and $\mathtt{grass}$ are known, but not the obstacles
it may encounter; 2) the Pac-Man has limited sensing capability, i.e.,
it can only detect obstacles, sense region labels, and collect rewards
within a local area around itself. The motion of the Pac-Man is modeled
by a weighted transition system $\mathcal{T}$ as in Def. \ref{def:WTS}
with four possible actions, \textquotedblleft up,\textquotedblright{}
\textquotedblleft down,\textquotedblright{} \textquotedblleft right,\textquotedblright{}
and \textquotedblleft left.\textquotedblright{} The timed temporal
task of Pac-Man is specified by an MITL formula $\phi=\phi_{h}\wedge\phi_{s}$,
where the hard constraints $\phi_{h}$ \textcolor{black}{enforce safety
requirement (e.g., }$\phi_{h}=\lnot\mathtt{obstacle}$\textcolor{black}{)
}that has to be fully satisfied while the soft constraints $\phi_{s}$
represent tasks that can be relaxed if
the environment does not permit (e.g., $\phi_{s}=\lnot\mathtt{grass}\land\lozenge_{t<10}\mathtt{pear}$).
\end{example}

In Example \ref{examp1}, the motion planning problem is challenging
since $\phi_{s}$ can be violated in multiple ways. For instance,
suppose that $\mathtt{grass}$ is in between $\mathtt{pear}$ and
Pac-Man, and it takes more than 10 seconds to reach $\mathtt{pear}$ if Pac-Man circumvents $\mathtt{grass}$. 
In this case, Pac-Man can either violate
the mission $\lnot\mathtt{grass}$ by traversing $\mathtt{grass}$
or violate the time constraints $\lozenge_{t<10}\mathtt{pear}$ by taking a longer but safer path.

To consider potentially infeasible specifications, we define 
the total violation cost of an MITL formula as follows.

\begin{defn}
Given a time run $\boldsymbol{r}=(q_{0},\tau_{0})\ldots(q_{n},\tau_{n})$ of a WTS $\mathcal{T}$, the total violation cost of an MITL formula $\phi$ is defined as
\begin{equation}
\mathcal{W}(\boldsymbol{r},\phi)=\stackrel[k=0]{n-1}{\sum}\omega(q_{k},q_{k+1})\omega_{v}(q_{k},q_{k+1},\phi),
\label{eq:total_cost}
\end{equation}
where $\omega(q_{k},q_{k+1})=\tau_{k+1}-\tau_{k}$ is the time required for the transition $(q_{k},q_{k+1})$ and $\omega_{v}(q_{k},q_{k+1},\phi)$ is defined as the violation cost of the transition  with respect to $\phi$. 
Then, the formal statement of the problem is expressed as follows.
\end{defn}

\begin{problem}
\label{prob1}Given a weighted transition system $\mathcal{T}$, and
an MITL formula $\phi=\phi_{h}\land\phi_{s}$, the control objective
is to design a multi-goal online planning strategy, in decreasing
order of priority, with  which 1) $\phi_{h}$ is fully satisfied; 2) $\phi_{s}$
is fulfilled as much as possible if $\phi_{s}$ is not feasible i.e. minimize the total violation cost $\mathcal{W}(\boldsymbol{r},\phi_{s})$; and
3)  the agent collects rewards as much as possible over an infinite horizon task operation.
\end{problem}

\section{Relaxed Automaton }

Sec. \ref{subsec:Relax} presents the procedure
of constructing the relaxed TBA to allow motion revision. Sec. \ref{subsec:Energy}
presents the design of energy function that guides the satisfaction
of MITL specifications. Sec. \ref{subsec: Update}
gives the online update of environment knowledge for motion planning.

\subsection{Relaxed Timed B\"uchi Automaton \label{subsec:Relax}}

% \begin{algorithm}
% \caption{\label{alg:Construct S,F}Construct set of states $\hat{S}$, initial
% states $\hat{S}_{0}$ and accepting states $\hat{F}$ of a relaxed
% TBA}

% \small

% \singlespacing

% \begin{algorithmic}[1]

% \Procedure {Input: } {MITL specification $\phi=\phi_{h}\land\phi_{s}$
% }

% {Output: } { $\hat{S},\hat{S}_{0},\hat{F}$}

% \State$\varPhi_{s}=\{\phi_{i}:\phi_{s}=\bigwedge_{i}\phi_{i}\}$

% \For{ $\phi_{i}\in\varPhi_{s}$ }

% \If{$\phi_{i}$ is temporally bounded }

% \State $\varphi_{i}=\{\phi_{i}^{sat},\phi_{i}^{vio},\phi_{i}^{unc}\}$;

% \ElsIf{$\phi_{i}$ is Type I }

% \State $\varphi_{i}=\{\phi_{i}^{sat},\phi_{i}^{unc}\}$;

% \Else{ $\varphi_{i}=\{\phi_{i}^{vio},\phi_{i}^{unc}\};$}

% \EndIf

% \EndFor

% \State$\varPsi_{s}=\Pi_{i}\varphi_{i}$;

% \State$\hat{S}=\{s_{i}:i=0,\ldots,n+1\},$where $n$ is the number
% of $\psi_{s}\in\varPsi_{s}$;

% \State$\hat{S}_{0}=s_{0},$where $s_{0}$ corresponds to $\psi_{0}=\bigwedge_{i}\phi_{i}^{unc}$;

% \State$\hat{F}=s_{F}$,where $s_{F}$ corresponds to $\psi_{F}=\bigwedge_{i\in I}\phi_{i}^{sat}\cap\bigwedge_{j\in J}\phi_{j}^{unc},$where
% $i\in I$ are the indexes of subformulas of $\phi_{s}$ that are
% either temporally bounded or of Type I, and $j\in J$ are the indexes
% of subformulas that are of Type II;

% \EndProcedure

% \end{algorithmic}
% \end{algorithm}
\begin{algorithm}

\caption{\label{alg:Construct S,F}Construct set of states $\hat{S}$, initial
states $\hat{S}_{0}$ and accepting states $\hat{F}$ of a relaxed
TBA}

\small
\singlespacing

\begin{algorithmic}[1]

\Procedure {Input: } {MITL specification $\phi=\phi_{h}\land\phi_{s}$
}

{Output: } { $\hat{S},\hat{S}_{0},\hat{F}$}
\State{ construct state relevant to $\phi_{h}$:}
\State{ add a state $\hat{s}_{sink}$}
\State{ construct states relevant to $\phi_{s}$:}
\State{ $\varPhi_{s}=\{\phi_{i}:\phi_{s}=\bigwedge_{i}\phi_{i}\}$}

\For{ $\phi_{i}\in\varPhi_{s}$ }

\If{$\phi_{i}$ is temporally bounded }

\State $\varphi_{i}=\{\phi_{i}^{sat},\phi_{i}^{vio},\phi_{i}^{unc}\}$;

\ElsIf{$\phi_{i}$ is non-temporally bounded of Type I }

\State $\varphi_{i}=\{\phi_{i}^{sat},\phi_{i}^{unc}\}$;

\Else{ $\varphi_{i}=\{\phi_{i}^{vio},\phi_{i}^{unc}\};$}

\EndIf

\EndFor

% \State$\varPsi_{s}=\Pi_{i}\varphi_{i}$, where $n$ is the number
% of $\psi_{s}\in\varPsi_{s}$;

\State $\psi_{s}^{j}=\underset{i}{\bigwedge}\phi_{i}^{state},\phi_{i}^{state}\in\varphi_{i}$;
\State $\varPsi_{s}=\left\{ \psi_{s}^{j}:j=0,1\ldots,n-1\right\} $ with $n=\prod_{i}\left|\varphi_{i}\right|$;
\State $\hat{S}=\{\hat{s}_{k}:k=0,1\ldots,n\}$;

\State$\hat{S}_{0}=\hat{s}_{0},$where $\hat{s}_{0}$ corresponds to $\psi_{s}^{0}=\bigwedge_{i}\phi_{i}^{unc}$;

\State$\hat{F}=\hat{s}_{F}$, where $\hat{s}_{F}$ corresponds to $\psi_{s}^{F}=\bigwedge_{i_{1}\in I_{1}}\phi_{i_{1}}^{sat}\cap\bigwedge_{i_{2}\in I_{2}}\phi_{i_{2}}^{unc}$,where
$i_{1}\in I_{1}$ are the indexes of sub-formulas of $\phi_{s}$ that are
either temporally bounded or of Type I, and $i_{2}\in I_{2}$ are the indexes
of sub-formulas that are of Type II;

\EndProcedure

\end{algorithmic}

\end{algorithm}

To address the violation of MITL tasks, the relaxed TBA is defined to contain two extra components (i.e., a continuous violation cost and a discrete violation cost) compared with the original TBA.
This section presents the procedure of constructing
a relaxed TBA for an MITL formula $\phi=\phi_{h}\land\phi_{s}$. 

First, we explain 
how to build the set of states in a relaxed TBA (see Alg.\ref{alg:Construct S,F}). Given the hard constraints $\phi_{h}$, which have to be fully satisfied and cannot be violated at any time, we add a sink state $\hat{s}_{sink}$ in the relaxed TBA to indicate the violation of hard constraints. 
% As for the soft constraints $\phi_{s}=\bigwedge_{i\in{1,...,n}}\phi_{i}$,
% an evaluation set $\varphi_{i}$ of a subformula $\phi_{i}$ is defined
% as

Before developing soft constraints $\phi_{s}$, a more detailed classification of temporal operators for MITL formulas is introduced.
An MITL specification $\phi$ can be written as $\phi=\bigwedge_{i\in{1,2,...,n}}\phi_{i}$
s.t. $\phi_{i}\neq\phi_{j},\forall i\neq j$. For each sub-formula $\phi_{i}$, if it is
temporally bounded, $\phi_{i}$ can be either satisfied, violated,
or uncertain \cite{Andersson2018}. If $\phi_{i}$ is non-temporally bounded, it can be
either satisfied/uncertain or violated/uncertain. 
Specifically,
a non-temporally bounded formula $\phi_{i}$ is of $\Type$
\mbox{I} (i.e., satisfied/uncertain) if $\phi_{i}$ cannot be concluded to be violated at any time
during a run since there remains a possibility for it to be satisfied
in the future. In contrast, it is of $\Type$ \mbox{II} (i.e., violated/uncertain) if $\phi_{i}$ cannot be concluded
to be satisfied during a run, since it remains possible to be violated in the future.
For instance, when $b=\infty$, $\diamondsuit_{[a,b]}$ is of $\Type$ \mbox{I} and $\Square_{[a,b]}$ is of $\Type$ \mbox{II}. The operator $\mathcal{U}_{[a,b]}$ is special since it results in two parts of semantics, which can be classified as $\Type$ \mbox{I} and \mbox{II}, respectively. Hence we treat formulas like $A\mathcal{U}_{\left[a,b\right]}B$ as a combination of two non-temporally bounded sub-formulas.

Based on above statement, for the soft constraints $\phi_{s}=\bigwedge_{i\in{1,...,n}}\phi_{i}$,
an evaluation set $\varphi_{i}$ of a sub-formula $\phi_{i}$ which represent possible satisfaction for a sub-formula is defined as

% \textcolor{blue}{As for the soft constraints $\phi_{s}=\bigwedge_{i\in{1,...,n}}\phi_{i}$,
% an evaluation set $\varphi_{i}$ of a subformula $\phi_{i}$ which represent possible satisfaction for a subformula based on an ongoing timed run is defined as}
\begin{equation}
\varphi_{i}=\left\{ \begin{array}{ll}
\left\{ \phi_{i}^{vio},\phi_{i}^{sat},\phi_{i}^{unc}\right\} , & \textrm{if \ensuremath{\phi_{i}}}\textrm{ is temporally bounded},\\
\left\{ \phi_{i}^{sat},\phi_{i}^{unc}\right\} , & \textrm{if \ensuremath{\phi_{i}}}\textrm{ is non-temporally }\\
 & \textrm{bounded of \ensuremath{\Type\ }\mbox{I} },\\
\left\{ \phi_{i}^{vio},\phi_{i}^{unc}\right\} , & \textrm{if \ensuremath{\phi_{i}}}\textrm{ is non-temporally }\\
 & \textrm{bounded of \ensuremath{\Type\ }\mbox{II}},
\end{array}\right.\label{eq:EvaluFcn_i}
\end{equation}
% Based on (\ref{eq:EvaluFcn_i}), let $\psi_{s}=\underset{i}{\bigwedge}\phi_{i}^{state}$ denote a possible outcome of the formula $\phi_{s}$ where
% $\phi_{i}^{state}\in\varphi_{i}$ denotes a possible evaluation of subformula
% $\phi_{i}$, and $\Psi_{s}$ denote the set of all possible evaluations of
%  $\phi_{s}$. 
% where $\ensuremath{\Type\ }\mbox{I}$ and  $\ensuremath{\Type\ }\mbox{II}$ are defined in Section \ref{sec:MITL}. 
Based on (\ref{eq:EvaluFcn_i}), a subformula evaluation $\psi_{s}$ of $\phi_{s}$
is defined as
\begin{equation}
\psi_{s}^{j}=\underset{i}{\bigwedge}\phi_{i}^{state},\phi_{i}^{state}\in\varphi_{i}
\label{eq:psi_s}.
\end{equation}
In (\ref{eq:psi_s}), $\psi_{s}^{j}$ represents one possible outcome of the formula, which can be obtained by taking an element from the evaluation set $\varphi_{i}$ for each sub-formula $\phi_i$, and then operating the conjunction of all these elements.
Each different combination corresponds to a sub-formula evaluation $\psi_s^{j}$. Let $\varPsi_{s}$ denote the
set of all sub-formula evaluations $\psi_{s}^{j}$ of $\phi_{s}$, the number of $\psi_{s}^{j}\in\varPsi_{s}$ is equal to the product of the number of elements in the evaluation set $\varphi_{i}$, which can be defined as $\varPsi_{s}=\left\{ \psi_{s}^{j},j=0,1\ldots,n-1\right\} $ with $n=\prod_{i}\left|\varphi_{i}\right|$ where $\left|\varphi_{i}\right|$ represents the number of elements in set $\varphi_{i}$. The set $\varPsi_{s}$ represents all possible outcomes of $\phi_{s}$ at any time.
Every possible $\psi_{s}^{j}\in\varPsi_{s}$ is associated with a state $\hat{s}$.
The initial state $\hat{s}_{0}$ is the state whose corresponding
sub-formulas are uncertain, which indicates no progress
has been made. The accepting state $\hat{s}_{F}$ is the state whose
corresponding temporally bounded sub-formulas and non-temporally bounded
sub-formulas of $\Type$ \mbox{I} are satisfied, while all non-temporally
bounded sub-formulas of $\Type$ \mbox{II} are uncertain.

The construction of the set of atomic propositions $\mathcal{AP}$, labeling function $\mathcal{L}$,
clocks $X$ and the map from states to clock constraints $I_{X}$
in relaxed TBA is the same as in TBA. Here consider two different types of violation cost, i.e.,
a state $\hat{s}\neq \hat{s}_{sink}$ can violate soft constraints $\phi_{s}$ by either continuous violation
(e.g., violating time constraints) or discrete violation (e.g., visiting
risky regions). To measure their violation degrees, the
outputs of continuous violation cost $v_{c}(\hat{s})$ and discrete violation
cost $v_{d}(\hat{s})$ for each state $\hat{s}\neq \hat{s}_{sink}$ are defined, respectively, as
% \begin{equation}
% v_{c}(s)=\left\{ \begin{array}{ll}
% k, & \textrm{if \ensuremath{k} temporally bounded operators}\\
%  & \textrm{are violated in state \ensuremath{s},}\\
% 0, & \textrm{otherwise.}
% \end{array}\right.
% \end{equation}
% \begin{equation}
% v_{d}(s)=\left\{ \begin{array}{ll}
% 0, & \textrm{if no non-temporally bounded }\\
%  & \textrm{operators are violated in state \ensuremath{s},}\\
% 1, & \textrm{otherwise.}
% \end{array}\right.
% \end{equation}
% \begin{equation}
% v_{c}(s)=\left\{ \begin{array}{ll}
% k, & \textrm{\ensuremath{\textrm{if}\ \exists\ \phi_{i}^{vio}\in\psi_{s}} that is temporally bounded and}\\
%  & \textrm{\ensuremath{k} is the number of violated time constraints}\\
% 0, & \textrm{otherwise}
% \end{array}\right.
% \end{equation}
% \begin{equation}
% v_{d}(s)=\left\{ \begin{array}{ll}
% 1, & \textrm{\ensuremath{\textrm{if}\ \exists\ \phi_{i}^{vio}\in\psi_{s}} that is non-temporally bounded }\\
% \\
% 0, & \textrm{otherwise}
% \end{array}\right.
% \end{equation}
\begin{equation}
v_{c}(\hat{s})=\left\{ \begin{array}{ll}
k, & \textrm{\textrm{if \ensuremath{\exists\phi_{1}^{vio},}\ensuremath{\phi_{2}^{vio}}, \ensuremath{\ldots\phi_{k}^{vio}\in\psi_{s}^{j}} that is temporally } }\\
 & \textrm{bounded, }\\
0, & \textrm{otherwise,}
\end{array}\right.
\end{equation}
\begin{equation}
v_{d}(\hat{s})=\left\{ \begin{array}{ll}
1, & \textrm{\ensuremath{\textrm{if}\ \exists\phi_{i}^{vio}\in\psi_{s}^{j}} that is non-temporally bounded, }\\
\\
0, & \textrm{otherwise,}
\end{array}\right.
\end{equation}

At the sink state $\hat{s}_{sink}$, the continuous and discrete violation costs are defined as $v_{c}(\hat{s}_{sink})=v_{d}(\hat{s}_{sink})=\infty$.

The next step is to define violation-based edges connecting
states, and the following definitions and
notations are introduced. 

\begin{defn}
Given soft constraints $\phi_{s}$, the distance set between $\psi_{s}$
and $\psi_{s}^{\prime}$ is defined as $|\psi_{s}-\psi_{s}^{\prime}|=\{\phi_{i}:\phi_{i}^{state^{\prime}}\neq\phi_{i}^{state}\}$.
That is, it consists of all sub-formulas $\phi_{i}$ that are under different evaluations.
\end{defn} 

We use $(\psi_{s},g,a)\rightarrow\psi_{s}^{\prime}$ to denote that
all sub-formulas $\phi_{i}\in|\psi_{s}-\psi_{s}^{\prime}|$ are (i)
evaluated as uncertain in $\psi_{s}$ (i.e., $\phi_{i}^{unc}\in\psi_{s}$)
and (ii) re-evaluated to be either satisfied or violated in $\psi_{s}^{\prime}$
(i.e., $\phi_{i}^{state^{\prime}}\in\psi_{s}^{\prime}$, where $state^{\prime}\in\left\{ vio,sat\right\} $)
if symbol $a$, which is read at time $t$, satisfies guard $g$.

\begin{figure*}
\centering{}\includegraphics[scale=0.25]{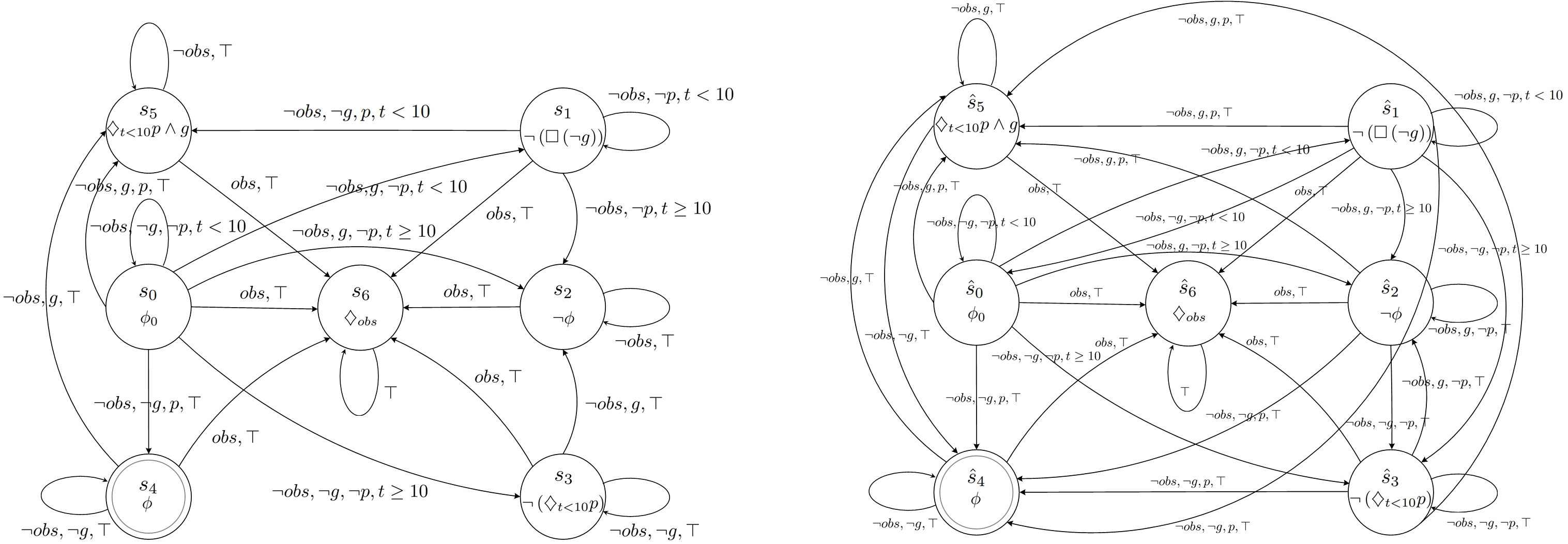}
%\centering{}\includegraphics[scale=0.5]{figure2_1.PNG}
\caption{\label{fig:relaxExample}(a) The TBA corresponding to $\phi=\phi_{h}\land\phi_{s}$,
where $\phi_{h}=\Square\lnot\mathtt{obs}$ and $\phi_{s}=\Square\lnot \mathtt{g}\land\lozenge_{t<10}\mathtt{p}$.
(b) The relaxed TBA corresponding to $\phi$.}
\end{figure*}

The edge construction can be summarized into four steps:

(1) Construct all edges corresponding to progress regarding the specifications (i.e., the edges that a TBA would have).

% (2) Construct edges corresponding to non-temporally bounded soft constraints that are no longer violated such that: $(\hat{s},g,a,\hat{s}^{\prime})\in\hat{E}$
% if (i) $\forall\phi_{i}\in|\psi_{s}-\psi_{s}^{\prime}|,\phi_{i}\in\phi_{s}$
% and is non-temporally bounded, and $\phi_{i}^{vio}\in\psi_{s}$, and
% (ii) $(\hat{s}^{\prime\prime},g,a,\hat{s}^{\prime})\in\hat{E}$ for
% some $\hat{s}^{\prime\prime}$ where $|\psi_{s}-\psi_{s}^{\prime}|=|\psi_{s}-\psi_{s}^{\prime\prime}|$
% or $(\hat{s}^{\prime},g,a^{\prime},\hat{s})\in\hat{E}$ where $a^{\prime}=2^{\mathcal{AP}}\setminus a$.
(2) Construct edges $\hat{E}$ of non-temporally bounded soft constraints that are no longer violated, such that $(\hat{s},g,a,\hat{s}^{\prime})\in\hat{E}$
satisfying all of the following conditions: (i) $\forall\phi_{i}\in|\psi_{s}-\psi_{s}^{\prime}|, \phi_{i}^{vio}\in\psi_{s}$ where
$\hat{s}$ corresponds to $\psi_{s}$, $\hat{s}^{\prime}$ corresponds
to $\psi_{s}^{\prime}$ and $\phi_{i}$ is non-temporally bounded, and
(ii) $(\hat{s}^{\prime\prime},g,a,\hat{s}^{\prime})\in\hat{E}$ for
some $\hat{s}^{\prime\prime}$ where $|\psi_{s}-\psi_{s}^{\prime}|=|\psi_{s}-\psi_{s}^{\prime\prime}|$
or $(\hat{s}^{\prime},g,a^{\prime},\hat{s})\in\hat{E}$ where $a^{\prime}=2^{\mathcal{AP}}\setminus a$.

% (3) Construct edges corresponding to temporally-bounded soft constraints that are no longer violated such that: $(\hat{s},g,a,\hat{s}^{\prime})\in\hat{E}$
% if (i) $\exists\phi_{i}\in|\psi_{s}-\psi_{s}^{\prime}|,\phi_{i}\in\phi_{s}$
% and is temporally bounded, and $\phi_{i}^{vio}\in\psi_{s}$, $\phi_{i}^{sat}\in\psi_{s}^{\prime}$, $\phi_{i}^{unc}\in\psi_{s}^{\prime\prime}$,
% where $(\hat{s}^{\prime\prime},g^{\prime},a,\hat{s}^{\prime})\in\hat{E}$
% and $(\hat{s}^{\prime\prime},g,a,\hat{s})\in\hat{E}$, (ii) $g=g^{\prime}\setminus\Phi(X_{i})$, where $X_{i}$ is the set of clocks associated with $\phi_{i}$, s.t.
% $\phi_{i}^{unc}\in\psi_{s}^{\prime}$ and $\phi_{i}^{vio}\in\psi_{s}$,
% and (iii) there exists no $\phi_{i}\in|\psi_{s}-\psi_{s}^{\prime}|$ s.t. $\phi_{i}\in\phi_{h}$.
(3) Construct edges $\hat{E}$ of temporally bounded soft constraints that are no longer violated, such that $(\hat{s},g,a,\hat{s}^{\prime})\in\hat{E}$ satisfying all the following conditions: (i) $\exists\phi_{i}\in|\psi_{s}-\psi_{s}^{\prime}|$, $\phi_{i}^{vio}\in\psi_{s}$, $\phi_{i}^{sat}\in\psi_{s}^{\prime}$, $\phi_{i}^{unc}\in\psi_{s}^{\prime\prime}$ where
$\hat{s}$ corresponds to $\psi_{s}$, $\hat{s}^{\prime}$ corresponds
to $\psi_{s}^{\prime}$, $\hat{s}^{\prime\prime}$ corresponds to $\psi_{s}^{\prime\prime}$
and $\phi_{i}$ is temporally bounded, (ii)
$(\hat{s}^{\prime\prime},g^{\prime},a,\hat{s}^{\prime})\in\hat{E}$
, $(\hat{s}^{\prime\prime},g,a,\hat{s})\in\hat{E}$ and $g=g^{\prime}\setminus\Phi(X_{i})$, where $X_{i}$ is the set of clocks associated with $\phi_{i}$, s.t.
$\phi_{i}^{unc}\in\psi_{s}^{\prime\prime}$ and $\phi_{i}^{vio}\in\psi_{s}$.
% (4) Construct self-loops such that $(\hat{s},g,a,\hat{s})\in\hat{E}$
% if $\exists\ (g,a)$ s.t. $g\subseteq g^{\prime}$ , $a\subseteq a^{\prime}$
% where $(\hat{s}^{\prime},g^{\prime},a^{\prime},\hat{s})\in\hat{E}$
% for some $\hat{s}^{\prime}$ and $(\hat{s},g^{\prime},a^{\prime},\hat{s}^{\prime\prime})\in\hat{E}$.

(4) Construct self-loops such that $(\hat{s},g,a,\hat{s})\in\hat{E}$
if $\exists\ (g,a)$ s.t. $g\subseteq g^{\prime}$ , $a\subseteq a^{\prime}$
where $(\hat{s}^{\prime},g^{\prime},a^{\prime},\hat{s})\in\hat{E}$
for some $\hat{s}^{\prime}$ and $(\hat{s},g^{\prime},a^{\prime},\hat{s}^{\prime\prime})\notin\hat{E}$
for any $\hat{s}^{\prime\prime}$.

In the first step, the edges of the original TBA are constructed except self-loops,
i.e., transitions from and to the same state. Then, we construct edges
from states where $v_{d}=1$, i.e., states corresponding to discrete
violation (step 2). These edges can be considered as alternative routes
to the ones in step 1, where some non-temporally bounded sub-formula/formulas
are violated at some points. Similarly, we construct edges from states
with $v_{c}>0$, i.e., states corresponding to continuous violations
(step 3). This ensures that the accepting states can be reached
when the time-bound action finally occurs, even after the deadline
is exceeded. Finally, we consider self-loops to ensure no deadlocks
in the automaton except the sink state $\hat{s}_{sink}$. Compared with TBA, the relaxed TBA allows more
transitions and enables task relaxation when $\phi_{s}$ is not fully
feasible.

\begin{defn}
An automata timed run $\boldsymbol{r}_{\mathcal{\hat{A}}}=(\hat{s}_{0},\tau_{0})\ldots(\hat{s}_{n},\tau_{n})$ of a relaxed TBA $\mathcal{\hat{A}}$, corresponding to the timed run $\boldsymbol{r}=(q_{0},\tau_{0})\ldots(q_{n},\tau_{n})$ is a sequence where $\hat{s}_{0}\in \hat{S}_{0}$, $\hat{s}_{j}\in \hat{S}$, and $(\hat{s}_{j},g_{j},a_{j},\hat{s}_{j+1})\in \hat{E}\ \forall j\geq0$
such that i) $\tau_{j}\models g_{j},j\geq0,$ and ii) $L(q_{j})\subseteq\mathcal{L}(\hat{s}_{j}),\forall j$. The continuous violation cost for the automata timed run is $\stackrel[k=0]{n-1}{\sum}v_{c}(\hat{s}_{k+1})(\tau_{k+1}-\tau_{k})$ and similarly the discrete violation cost is $\stackrel[k=0]{n-1}{\sum}v_{d}(\hat{s}_{k+1})(\tau_{k+1}-\tau_{k})$. 

\label{def:auto_timedrun}
\end{defn}

\begin{example}
As a running example in Fig. \ref{fig:relaxExample}.
Consider an MITL specification $\phi=\phi_{h}\land\phi_{s}$ with
$\phi_{h}=\Square\lnot\mathtt{obs}$ and $\phi_{s}=\Square\lnot \mathtt{g}\land\lozenge_{t<10}\mathtt{p}$,
where $\mathtt{obs}$ represents obstacles, and $\mathtt{g}$ and $\mathtt{p}$ represent
the grass and pear, respectively. The TBA and the corresponding relaxed
TBA are shown in Fig. \ref{fig:relaxExample}. 
% The soft constraint
% $\phi_{s}$ is composed of two subformulas: $\phi_{1}=\lnot \mathtt{g}$ and
% $\phi_{2}=\lozenge_{t<10}\mathtt{p}$, where $\phi_{1}$ is non-temporally
% bounded $\Type$ \mbox{II} and $\phi_{2}$ is temporally bounded.
% Hence $\phi_{1}$ can be evaluated as violated or uncertain while
% $\phi_{2}$ can be evaluated as violated, uncertain or satisfied.
% Following Alg. \ref{alg:Construct S,F}, the relaxed TBA has 7 states,
% which satisfy the hard constraints except that $s_{7}$ is a sink state indicating that the hard constraint $\phi_{h}$ is violated.
% For $\phi_{s},$ the initial state $s_{1}\sim\phi_{1}^{unc}\text{\ensuremath{\bigwedge}}\phi_{2}^{unc}$
% corresponds to subformulas evaluated as uncertain. The accepting state
% $s_{5}\sim\phi_{1}^{unc}\bigwedge\phi_{2}^{sat}$ corresponds to $\phi_{1}$
% evaluated as uncertain and $\phi_{2}$ as satisfied. For the rest
% of the states, we denote by $s_{2}\sim\phi_{1}^{vio}\bigwedge\phi_{2}^{unc}$,
% $s_{3}\sim\phi_{1}^{vio}\bigwedge\phi_{2}^{vio}$, $s_{4}\sim\phi_{1}^{unc}\bigwedge\phi_{2}^{vio}$,
% $s_{6}\sim\phi_{1}^{vio}\bigwedge\phi_{2}^{sat}$. 
% There are two clock constraints in this example: $t<10$ associated with
% states corresponding to $\phi_{2}^{sat}$, and $t\geq10$ 
% associated with $\phi_{2}^{vio}$. The first clock constraint is then
% mapped to $s_{5}$ and $s_{6}$, and the second to $s_{3}$ and $s_{4}$.
% The continuous and discrete violation costs are mapped such that $v_{c}(\hat{S})=[0\ 0\ 1\ 1\ 0\ 0\ \infty]$
% and $v_{d}(\hat{S})=[0\ 1\ 1\ 0\ 0\ 1\ \infty]$.
The soft constraint
$\phi_{s}$ is composed of two subformulas: $\phi_{1}=\Square\lnot \mathtt{g}$ and
$\phi_{2}=\lozenge_{t<10}\mathtt{p}$, where $\phi_{1}$ is non-temporally
bounded of $\Type$ \mbox{II} and $\phi_{2}$ is temporally bounded.
Hence $\phi_{1}$ can be evaluated as violated or uncertain while
$\phi_{2}$ can be evaluated as violated, uncertain or satisfied, i.e. the corresponding evaluation sets are $\varphi_{1}=\left\{ \phi_{1}^{unc},\phi_{1}^{vio}\right\} $ and $\varphi_{2}=\left\{ \phi_{2}^{unc},\phi_{2}^{vio},\phi_{2}^{sat}\right\} $, respectively. By operating the conjunction of the first element in set $\varphi_{1}$ and set $\varphi_{2}$, a sub-formula evaluation $\psi_{s}^{0}=\phi_{1}^{unc}\land\phi_{2}^{unc}$ is obtained. Similarly, we can enumerate all sub-formula evaluations. Therefore, the set of all sub-formula evaluations of the formula $\phi_{s}$ is 
$\varPsi_{s}=\left\{ \phi_{1}^{unc}\wedge\phi_{2}^{unc},\phi_{1}^{vio}\wedge\phi_{2}^{unc},\phi_{1}^{vio}\wedge\phi_{2}^{vio},\phi_{1}^{unc}\wedge\phi_{2}^{vio},\right.$ $\left.\phi_{1}^{unc}\wedge\phi_{2}^{sat},\phi_{1}^{vio}\wedge\phi_{2}^{sat}\right\} $
with $\psi_{s}^{j}\in \varPsi_{s}$.
Following Alg. \ref{alg:Construct S,F}, the relaxed TBA has 7 states,
which satisfy the hard constraints except that $\hat{s}_{6}$ is a sink state indicating that the hard constraint $\phi_{h}$ is violated.
For $\phi_{s},$ the initial state $\hat{s}_{0}\sim\phi_{1}^{unc}\wedge\phi_{2}^{unc}$
corresponds to sub-formulas evaluated as uncertain. The accepting state
$\hat{s}_{4}\sim\phi_{1}^{unc}\wedge\phi_{2}^{sat}$ corresponds to $\phi_{1}$
evaluated as uncertain and $\phi_{2}$ as satisfied. For the rest
of the states, we denote by $\hat{s}_{1}\sim\phi_{1}^{vio}\wedge\phi_{2}^{unc}$,
$\hat{s}_{2}\sim\phi_{1}^{vio}\wedge\phi_{2}^{vio}$, $\hat{s}_{3}\sim\phi_{1}^{unc}\wedge\phi_{2}^{vio}$,
$\hat{s}_{5}\sim\phi_{1}^{vio}\wedge\phi_{2}^{sat}$. 
There are two clock constraints in this example: $t<10$ associated with
states corresponding to $\phi_{2}^{sat}$, and $t\geq10$ 
associated with $\phi_{2}^{vio}$. The first clock constraint is then
mapped to $\hat{s}_{4}$ and $\hat{s}_{5}$, and the second to $\hat{s}_{2}$ and $\hat{s}_{3}$.
The continuous and discrete violation costs are mapped such that $v_{c}(\hat{S})=[0\ 0\ 1\ 1\ 0\ 0\ \infty]$
and $v_{d}(\hat{S})=[0\ 1\ 1\ 0\ 0\ 1\ \infty]$. 
\end{example}

Compared with TBA, the relaxed TBA allows more
transitions, enables task relaxation and measure its violation when $\phi_{s}$ is not fully
feasible. Since a traditional product automaton $\mathcal{P}=\mathcal{T\times A}$ cannot handle the infeasible case, a relaxed product automaton is introduced as follow. 

\begin{defn}
\label{def:relaxed product automaton}Given a  WTS $\mathcal{T}=\textrm{\ensuremath{\left(Q,q_{0},\delta,\mathcal{AP},L,\mathcal{\omega}\right)}}$
and a relaxed TBA $\mathcal{\hat{A}}=(\hat{S},\hat{S}_{0},\mathcal{AP},\mathcal{L},X,I_{X},v_{c},v_{d},\hat{E},\hat{F})$,
the relaxed product automaton (RPA) $\hat{\mathcal{P}}=\mathcal{T\times\hat{A}}$
is defined as a tuple $\hat{\mathcal{P}}=\{\hat{P},\hat{P_{0}},\mathcal{AP},L_{\mathcal{\hat{P}}},\delta_{\mathcal{\hat{P}}},I_{X}^{\hat{\mathcal{P}}},v_{c}^{\hat{\mathcal{P}}},v_{d}^{\hat{\mathcal{P}}},\mathcal{F_{\hat{P}}},\omega_{\mathcal{\hat{P}}}\}$,
$\hat{P}\subseteq\{(q,\hat{s})\in Q\times\hat{S}:L(q)\subseteq\mathcal{L}(\hat{s})\}$
is the set of states; $\hat{P_{0}}=\{q_{0}\}\times\hat{S}_{0}$
is the set of initial states; $L_{\mathcal{\hat{P}}}=\hat{P}\rightarrow2^{\mathcal{AP}}$
is a labeling function, i.e., $L_{\mathcal{\hat{P}}}(\hat{p})=L(q)$;
$\delta_{\mathcal{\hat{P}}}\subseteq\hat{P}\times\hat{P}$ is the
set of transitions defined such that $((q,\hat{s}),(q^{\prime},\hat{s}^{\prime}))\in\delta_{\mathcal{\hat{P}}}$
if and only if ($q,q^{\prime})\in\delta$ and $\exists g,a,\textrm{s.t.\ }(\hat{s},g,a,\hat{s}^{\prime})\in\hat{E}$; $I_{X}^{\hat{\mathcal{P}}}(\hat{p})=I_{X}(\hat{s})$
is a map of clock constraints; $v_{c}^{\hat{\mathcal{P}}}(\hat{p})=v_{c}(\hat{s})$ is
the continuous violation cost; $v_{d}^{\hat{\mathcal{P}}}(\hat{p})=v_{d}(\hat{s})$ is
the discrete violation cost; $\mathcal{F}_{\mathcal{\hat{P}}}=Q\times\hat{F}$
are accepting states; $\omega_{\mathcal{\hat{P}}}\colon\delta_{\mathcal{\hat{P}}}\rightarrow\mathbb{R}^{+}$
is the positive weight function, i.e., $\omega_{\mathcal{\hat{P}}}(\hat{p},\hat{p}^{\prime})=\omega(q,q^{\prime})$.
\label{def:RPA}
\end{defn}

By accounting continuous and discrete violation simultaneously, the violation cost with respect to $\phi_s$ is defined as 
\begin{equation}
\omega_{v}^{\hat{\mathcal{P}}}(\hat{p}_{k},\hat{p}_{k+1},\phi_{s})=(1-\alpha)v_{c}^{\hat{\mathcal{P}}}(\hat{p}_{k+1})+\alpha v_{d}^{\hat{\mathcal{P}}}(\hat{p}_{k+1}),
\end{equation}
where $\alpha\in[0,1]$ measures the relative importance between
continuous and discrete violations. Then based on $\mathcal{W}(\boldsymbol{r},\phi)$ defined in (\ref{eq:total_cost}), the total weight of a path $\hat{\boldsymbol{p}}=(q_{0},\hat{s}_{0})\ldots(q_{n},\hat{s}_{n})$
for $\hat{\mathcal{P}}$ is 
\begin{equation}
\mathcal{W}(\boldsymbol{\hat{p}})=\stackrel[k=0]{n-1}{\sum}\omega_{\mathcal{\hat{P}}}(\hat{p}_{k},\hat{p}_{k+1})\omega_{v}^{\hat{\mathcal{P}}}(\hat{p}_{k},\hat{p}_{k+1},\phi_{s}),
\label{eq:WeightFcn}
\end{equation}
where $\mathcal{W}(\boldsymbol{\hat{p}})$ measures the total violations with respect to $\phi_{s}$ in the WTS. Hence, by minimizing the violation
of $\phi_{s}$ a run  $\hat{\boldsymbol{p}}$ of $\hat{\mathcal{P}}$
can fulfill $\phi_{s}$ as much as possible.

\subsection{Energy Function \label{subsec:Energy}}

 Inspired by previous work \cite{Cai2020c}, we design a hybrid Lyapunov-like energy function consisting of different violation costs. Such a design can measure the minimum distance to the accepting sets from the current state and enforce the accepting condition by decreasing the energy as the system evolves.
 
Based on (\ref{eq:WeightFcn}), $d(\hat{p}_{i},\hat{p}_{j})=\textrm{min}_{\hat{\boldsymbol{p}}\in\mathcal{\mathcal{\hat{D}}}(\hat{p}_{i},\hat{p}_{j})}\mathcal{W}(\hat{\boldsymbol{p}})$ is the shortest path from $\hat{p}_{i}$ to $\hat{p}_{j}$ , where
$\mathcal{\hat{D}}(\hat{p}_{i},\hat{p}_{j})$ is the set of all possible paths.  

For $\hat{p}\in\hat{P}$, we design the energy function as
\begin{equation}
J(\hat{p})=\left\{ \begin{array}{cc}
\underset{\hat{p}^{\prime}\in\mathcal{F}^{\ast}}{\textrm{min}}d(\hat{p},\hat{p}^{\prime}), & \textrm{ if}\ \hat{p}\notin\mathcal{F}^{\ast},\\
0, & \textrm{ if\ }\hat{p}\in\mathcal{F}^{\ast},
\end{array}\right.
\label{eq:energy_fcn}
\end{equation}
where $\mathcal{F}^{\ast}$ is the largest self-reachable subset of
the accepting set $\mathcal{F}_{\hat{P}}$. Since $\omega_{\mathcal{\hat{P}}}$
is positive by definition, $d(\hat{p},\hat{p}^{\prime})>0$ for all
$\hat{p},\hat{p}^{\prime}\in\hat{P}$, which implies that $J(\hat{p})\geq0$.
Particularly, $J(\hat{p})=0$ if $\hat{p}\in\mathcal{F}^{\ast}$.
If a state in $\mathcal{F}^{\ast}$ is reachable from $\hat{p}$,
then $J(\hat{p})\neq\infty$, otherwise $J(\hat{p})=\infty$. Therefore,
$J(\hat{p})$ indicates the minimum distance from $\hat{p}$ to $\mathcal{F}^{\ast}$.

\begin{thm}

For the energy function designed in (\ref{eq:energy_fcn}), if a
trajectory $\boldsymbol{\hat{p}}=\hat{p}_{1}\hat{p}_{2}\ldots \hat{p}_{n}$ is accepting, there is no state $p_{i}$,$\forall i=1,2,\dots n$ with $J(\hat{p}_{i})=\infty$, and 
all accepting states in $\hat{p}$ are in the set $\mathcal{F}^{\ast}$ with energy 0. In addition, for any state
$\hat{p} \in \hat{P}$ with $\hat{p}\notin\mathcal{F}^{\ast}$ and $J(\hat{p})\neq\infty$, there exists at least one state $\hat{p}^{\prime}$ with $\hat{p},\hat{p}^{\prime}\in\delta_{\mathcal{\hat{P}}}$ such that $J(\hat{p}^{\prime})<J(\hat{p})$.

\label{thm:energy_fcn}
\end{thm}

\begin{IEEEproof}
Considering an accepting state $\hat{p}_{i}\in\mathcal{F}_{\mathcal{\hat{P}}}$. Suppose $\hat{p}_{i}\notin\mathcal{F}^{\ast}$. By definition \ref{def:RPA}, $\boldsymbol{\hat{p}}$ intersects $\mathcal{F}_{\mathcal{\hat{P}}}$ infinitely many times which indicates there exists another accepting state $\hat{p}_{j}\in\mathcal{F}_{\mathcal{\hat{P}}}$ reachable from $\hat{p}_{i}$. If $\hat{p}_{j}\in\mathcal{F}^{\ast}$, then by definition of $\mathcal{F}^{\ast}$, $\hat{p}_{i}$ must be in $\mathcal{F}^{\ast}$ which contradicts the assumption that 
$\hat{p}_{i}\notin\mathcal{F}^{\ast}$. For the case $\hat{p}_{j}\notin\mathcal{F}^{\ast}$, there must exist a non-trivial strongly connnected component(SCC) composed of accepting states reachable from $\hat{p}_{j}$. All states in SCC belong to $\mathcal{F}^{\ast}$. Since the SCC is reachable from $\hat{p}_{j}$, it implies $\hat{p}_{j}\in\mathcal{F}^{\ast}$, which contradicts the assumption. Thus all accepting states in $\hat{p}$ must be in $\mathcal{F}^{\ast}$ with energy zero based on (\ref{eq:energy_fcn}). Since $\mathcal{F}^{\ast}$ is reachable by any state in $\hat{p}$, $J(\hat{p}_{i})\neq\infty$, $\forall i=1,2,\dots n$.

If $J(\hat{p})\neq\infty$ for $\hat{p}\in \hat{P}$, (\ref{eq:energy_fcn}) indicates $\mathcal{F}^{\ast}$ is reachable from $\hat{p}$. Then there exists a shortest trajectory $\boldsymbol{\hat{p}}=\hat{p}_{1}\hat{p}_{2}\ldots \hat{p}_{n}$ where $\hat{p}_{1}=\hat{p}$ and $\hat{p}_{n}\in \mathcal{F}^{\ast}$. Bellman's optimal principle states that there exists a state $\hat{p}^{\prime}$ with $(\hat{p},\hat{p}^{\prime})\in\delta_{\mathcal{\hat{P}}}$ such that $J(\hat{p}^{\prime})<J(\hat{p})$.
\end{IEEEproof}
Theorem \ref{thm:energy_fcn} indicates that the generated path will eventually satisfy the acceptance condition of $\mathcal{\hat{P}}$ as long as the energy function keeps decreasing.

\subsection{Automaton Update \label{subsec: Update}}

% The system model needs to be updated
% based on the sensed information during the runtime to facilitate motion planning. The update procedure
% is outlined in Alg. \ref{alg:Automaton-Update}. Let $Q_{N}$ denote
% the set of sensible neighboring states and let $\left\llbracket \hat{p}\right\rrbracket =\{\hat{p}=(q,\hat{s})\mid q\in Q_{N}\}$
% denote a class of $\hat{p}$ sharing the same neighboring states.
% Specifically, we use $\textrm{Info}(\hat{p})=\{L_{\hat{\mathcal{P}}}(\hat{p}^{\prime})\mid\hat{p}^{\prime}\in\textrm{Sense}(\hat{p})\}$
% to denote the newly observed labels of $\hat{p}^{\prime}$ that are
% different from the current knowledge, where $\textrm{Sense}(\hat{p})$
% represents a local set of states that the agent at
% $\hat{p}$ can sense. If the sensed labels $L_{\hat{\mathcal{P}}}(\hat{p}^{\prime})$
% are consistent with the current knowledge of $\hat{p}^{\prime}$,
% $\Info(\hat{p})=\emptyset$; otherwise, the properties of $\hat{p}^{\prime}$
% have to be updated. 
% Let $\boldsymbol{J}(\left\llbracket \hat{p}\right\rrbracket )\in\mathbb{R}^{\left|\hat{p}\right|}$
% denote the stacked $J$ for all $\hat{p}\in\left\llbracket \hat{p}\right\rrbracket $.
% The terms $\mathit{\boldsymbol{J}}$ are initialized from the initial
% knowledge of the environment.
% At each step, if $\textrm{Info}(\hat{p})\neq\emptyset$, the weight $w$ and the energy
% function $J$ for each state of $\left\llbracket \hat{p}\right\rrbracket $
% are updated. 
% Note that the largest self-reachable set $\mathcal{F}^{\ast}$
% remains the same during the automaton update in Alg. \ref{alg:Automaton-Update}.

The system model needs to be updated
according to the sensed information during the runtime to facilitate motion planning. The update procedure
is outlined in Alg. \ref{alg:Automaton-Update}. Let $\textrm{Info}(\hat{p})=\{L_{\hat{\mathcal{P}}}(\hat{p}^{\prime})\mid\hat{p}^{\prime}\in\textrm{Sense}(\hat{p})\}$
denote the newly observed labels of $\hat{p}^{\prime}$ that are
different from the current knowledge, where $\textrm{Sense}(\hat{p})$
represents neighbor states that the agent at current state
$\hat{p}$ can detect and observe. Denote the sensing range is $N_s$. If the sensed labels $L_{\hat{\mathcal{P}}}(\hat{p}^{\prime})$
are consistent with the current knowledge of $\hat{p}^{\prime}$,
$\Info(\hat{p})=\emptyset$; otherwise, the properties of $\hat{p}^{\prime}$
have to be updated. 
Let $\boldsymbol{J}\in\mathbb{R}^{\left|\hat{P}\right|}$
denote the stacked $J$ for all $\hat{p}\in \hat{P}$.
The terms $\mathit{\boldsymbol{J}}$ are initialized from the initial
knowledge of the environment.
At each step, if $\textrm{Info}(\hat{p})\neq\emptyset$, the weight $\omega_{\mathcal{\hat{P}}}(\hat{p}^{\prime},\hat{p}^{\prime\prime})$ and $\omega_{\mathcal{\hat{P}}}(\hat{p}^{\prime\prime},\hat{p}^{\prime})$ for states that satisfy $\hat{p}^{\prime}\in\textrm{Sense}(\hat{p})$ and $(\hat{p}^{\prime},\hat{p}^{\prime\prime})\in\delta_{\mathcal{\hat{P}}}$ are updated. Then the energy
function $\boldsymbol{J}$
is updated.

\begin{lem}
\label{lem:self-set}
The largest self-reachable set $\mathcal{F}^{\ast}$ remains the same
during the automaton update in Alg. \ref{alg:Automaton-Update}.
\end{lem}

\begin{IEEEproof}
Given $\mathcal{\hat{P}}_{(\hat{p},\delta_{\hat{\mathcal{P}}})}$,
the graph induced from $\mathcal{\hat{P}}_{(\hat{p},\delta_{\hat{\mathcal{P}}})}$
by neglecting the weight of each transition is denoted by $\mathcal{G}(\hat{p},\delta_{\hat{\mathcal{P}}})$.
Similar to \cite{Cai2020c}, Alg. \ref{alg:Automaton-Update}. only updates the cost of each transition so that the topological structure of $\mathcal{G}(\hat{p},\delta_{\hat{\mathcal{P}}})$
and its corresponding $\mathcal{F}^{\ast}$ remain the same. 
\end{IEEEproof}

Lemma \ref{lem:self-set} indicates that $\mathcal{F}^{\ast}$ doesn't need to be updated whenever newly sensed information caused
by unknown obstacles is obtained. Therefore, it reduces the complexity. As a result, the $\mathcal{F}^{\ast}$ is computed off-line, and the construction of $\mathcal{F}^{\ast}$ involves the
computation of $d(\hat{p},\hat{p}^{\prime})$ for all $\hat{p}^{\prime}\in\mathcal{F_{\hat{P}}}$
and the check of terminal conditions \cite{Ding2014}.

\begin{algorithm}
\caption{\label{alg:Automaton-Update}Automaton Update}

\small

\singlespacing

\begin{algorithmic}[1]

\Procedure {Input: } {the current state $\hat{p}=(q,\hat{s}),$
the current $\boldsymbol{J},\mathcal{F}^{\ast}$and
$\textrm{Info}(\hat{p})$}

{Output: } { the updated $\boldsymbol{J}$}

\If { $\textrm{Info}(\hat{p})\neq\emptyset$}

\For{ all $\hat{p}^{\prime}=(q^{\prime},\hat{s}^{\prime})\in\textrm{Sense}(\hat{p})$
such that $L_{\hat{\mathcal{P}}}(\hat{p}^{\prime})\in\textrm{Info}(\hat{p})$}

\For{ all $\hat{p}^{\prime}$ such that $(\hat{p}^{\prime},\hat{p}^{\prime\prime})\in\delta_{\mathcal{\hat{P}}}$}

\State update the labels of $L_{\hat{\mathcal{P}}}(\hat{p}^{\prime})$
according to $L(q^{\prime})$;
\State update the weight $\omega_{\mathcal{\hat{P}}}(\hat{p}^{\prime},\hat{p}^{\prime\prime})$ and $\omega_{\mathcal{\hat{P}}}(\hat{p}^{\prime\prime},\hat{p}^{\prime})$;

\EndFor

\EndFor

\State update $\boldsymbol{J}$;

\EndIf

\EndProcedure

\end{algorithmic}
\end{algorithm}

\section{Control Synthesis of MITL Motion Planning}

\begin{figure*}[t]
\centering{}\includegraphics[scale=0.5]{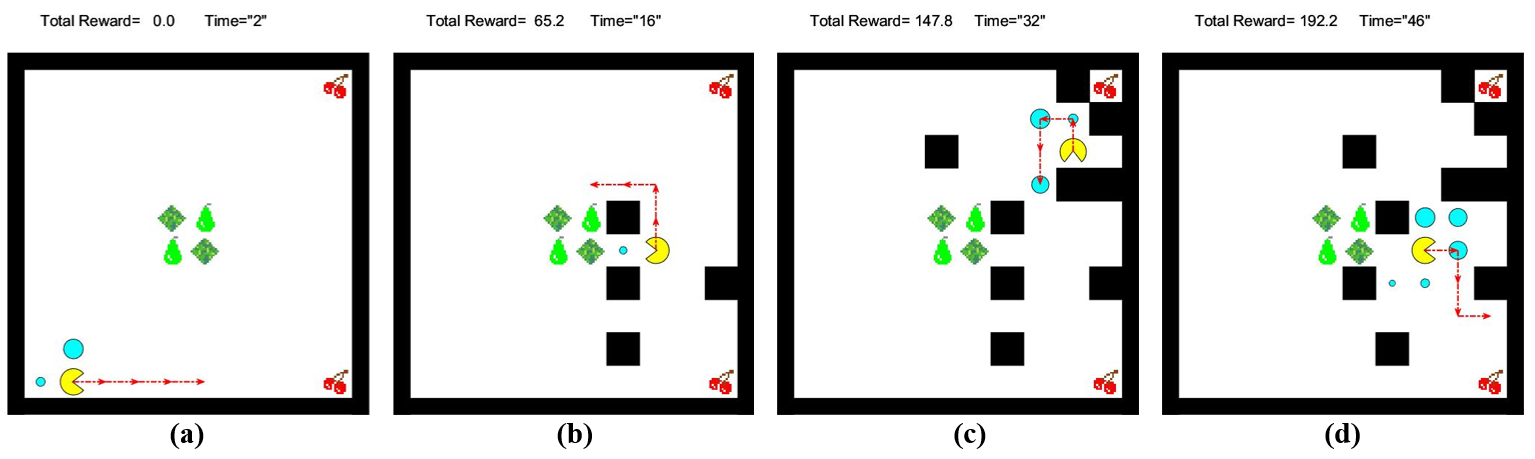}\caption{\label{fig:Snapshots}Snapshots of the motion planning. The red dotted
arrow line represents the predicted trajectory at the current time. Cyan dots represent the rewards with size proportional to
their value.
(a) Pac-Man plans to reach pear within the specified time interval.
(b) Since the soft task is infeasible, Pac-Man chooses to violate
the temporally bounded operators. (c) and (d) show when the desired
task is accomplished, Pac-Man revises its motion plan to go to the
cherry at the right bottom corner since the right top one is not accessible.}
\end{figure*}
The control synthesis of the MITL motion planning strategy is based
on receding horizon control (RHC). The idea of RHC is to solve an
online optimization problem by maximizing the utility function over
a finite horizon $N$ and produces a predicted optimal path
at each time step. With only the first predicted step applied, the
optimization problem is repeatedly solved to predict optimal paths.
Specifically, based on the current state $\hat{p}_{k}$, let $\hat{\boldsymbol{p}}_{k}=\hat{p}_{1\mid k}\hat{p}_{2\mid k}\ldots\hat{p}_{N\mid k}$
denote a predicted path of horizon $N$ at time $k$ starting
from $\hat{p}_{k}$, where $\hat{p}_{i\mid k}\in\hat{P}$ satisfies
$(\hat{p}_{i\mid k},\hat{p}_{i+1\mid k})\in\delta_{\hat{\mathcal{P}}}$
for all $i=1,...,N-1$, and $(\hat{p}_{k},\hat{p}_{1\mid k})\in\delta_{\mathcal{\hat{P}}}$.
Let $\textrm{Path}(\hat{p}_{k},N)$ be the set of paths of
horizon $N$ generated from $\hat{p}_{k}$. Note that a predicted
path $\hat{\boldsymbol{p}}_{k}\in\textrm{Path}(\hat{p}_{k},N)$
can uniquely project to a trajectory $\gamma_{\mathcal{T}}(\hat{\boldsymbol{p}}_{k})=\boldsymbol{q}=q_{1}\cdots q_{N}$
on $\mathcal{T}$, where $\gamma_{\mathcal{T}}(\hat{p}_{i\mid k})=q_{i}$,
$\forall i=1,\ldots,N$. The choice of the finite horizon $N$ depends
on the local sensing range $N_s$ of the agent. The total reward along the
predicted path $\hat{\boldsymbol{p}}_{k}$ is $\boldsymbol{R}(\gamma_{\mathcal{T}}(\hat{\boldsymbol{p}}_{k}))=\stackrel[i=1]{N}{\sum}R_{k}(\gamma_{\mathcal{T}}(\hat{p}_{i\mid k}))$.

Based on (\ref{eq:WeightFcn}), for every predicted path $\hat{\boldsymbol{p}}_{k}$ the total violation cost is $\mathcal{W}(\boldsymbol{\hat{p}_{k}})$.
Then the utility function of RHC is designed as
% \begin{equation}
% \mathbf{U}(\hat{\boldsymbol{p}_{k}})=\boldsymbol{R}(\gamma_{\mathcal{T}}(\hat{\boldsymbol{p}_{k}}))-\mathbf{V}_{c}^{\mathcal{P}}(\hat{\boldsymbol{p}_{k}})-\kappa\mathbf{V}_{d}^{\mathcal{P}}(\hat{\boldsymbol{p}_{k}}),
% \end{equation}
\begin{equation}
\mathbf{U}(\hat{\boldsymbol{p}}_{k})=\boldsymbol{R}(\gamma_{\mathcal{T}}(\hat{\boldsymbol{p}}_{k}))-\beta\mathcal{W}(\boldsymbol{\hat{p}_{k}}),
\end{equation}
where $\beta$ is the relative penalty. 
% The difference between $\beta_{1}$ and $\beta_{2}$ is that $\beta_{1}$ works on the weight function  while $\beta_{2}$ works  on  the  objective  function.
% where $\kappa\in\mathbb{R}^{+}$ is a tuning parameter indicating
% the relative importance between continuous and discrete violation
% cost.  

% If a larger $\kappa$ is applied, optimizing $\mathbf{U}(\hat{\boldsymbol{p}_{k}})$
% tends to bias the selection of paths which prefers avoiding discrete
% violation to continuous violation.
% \textcolor{blue}{(Mingyu: Need more detailed descriptions about the priorities of the objectives that matches the Problem 1. It should include discrete violation, continuous violation, reward)}
By applying large $\beta$, maximizing the utility $\mathbf{U}(\hat{\boldsymbol{p}}_{k})$ tends to bias the selection of paths
towards the objectives, in the decreasing order, of 1) hard constraints
$\phi_{h}$ satisfaction, 2) fulfilling soft constraints $\phi_{s}$ as much as possible,
and 3) collecting time-varying rewards as much as possible. Note that continuous and discrete violations are optimized simultaneously based on the preference weight $\alpha$ in $\mathcal{W}(\boldsymbol{\hat{p}_{k}})$.
To
satisfy the acceptance condition of $\mathcal{\hat{P}}$,
we consider the energy function-based constraints simultaneously.

The initial predicted path from $\hat{P_{0}}$ can be identified by solving
\begin{equation}
\begin{aligned}\hat{\boldsymbol{p}}_{0,opt}= & \underset{\hat{\boldsymbol{p}}_{0}\in\textrm{Path}(\hat{P}_{0},N)}{\textrm{argmax}}\mathbf{U}(\hat{\boldsymbol{p}}_{0}),\\
 & \textrm{ subject to: }J(\hat{p}_{0})<\infty.
\end{aligned}
\label{eq:P0}
\end{equation}
% The constraint $J(\hat{p}_{0})<\infty$ guarantees the existence of
% a satisfying path from $\hat{P_{0}}$ over $\mathcal{\hat{P}}$. 
The constraint $J(\hat{p}_{0})<\infty$ is critical because otherwise, the path starting from $\hat{p}_{0}$ cannot be accepting.

After determining the initial state $\hat{p}_{0}^{\ast}=\hat{p}_{1|0,opt}$,
where $\hat{p}_{1|0,opt}$ is the first element of $\hat{\boldsymbol{p}}_{0,opt}$
, RHC will be employed repeatedly to determine the optimal states
$\hat{p}_{k}^{\ast}$ for $k=1,2,\ldots$. At each time instant $k$,
a predicted optimal path $\hat{\boldsymbol{p}}_{k,opt}=\hat{p}_{1\mid k,opt}\hat{p}_{2\mid k,opt}\cdots\hat{p}_{N\mid k,opt}$
is constructed based on $\hat{p}_{k-1}^{\ast}$ and $\hat{\boldsymbol{p}}_{k-1,opt}$
obtained at time $k-1$. Note that only $\hat{p}_{1\mid k,opt}$ will
be applied at time $k$, i.e., $\hat{p}_{k}^{\ast}=\hat{p}_{1\mid k,opt}$,
which will then be used with $\hat{\boldsymbol{p}}_{k,opt}$ to generate
$\hat{\boldsymbol{p}}_{k+1,opt}$.
\begin{thm}
For each time k = 1, 2 . . ., provided $\hat{p}_{k-1}^{\ast}$ and
$\hat{\boldsymbol{p}}_{k-1,opt}$ from previous time step, consider
a RHC
\begin{equation}
\hat{\boldsymbol{p}}_{k,opt}=\underset{\hat{\boldsymbol{p}}_{k}\in\textrm{Path}(\hat{p}_{k-1}^{\ast},N)}{\textrm{argmax}}\ \ \mathbf{U}(\hat{\boldsymbol{p}}_{k}),
\label{eq:P_kopt}
\end{equation}
subject to the following constraints:
\begin{enumerate}
\item $J(\hat{p}_{N\mid k})<J(\hat{p}_{N\mid k-1,opt})$ if $J(\hat{p}_{k-1}^{\ast})>0$
and $J(\hat{p}_{i\mid k-1,opt})\neq0$ for all $i=1,\ldots,N$;
\item $J(\hat{p}_{i_{0}(\hat{p}_{k-1,opt})-1\mid k})=0$ if $J(\hat{p}_{k-1}^{\ast})>0$
and $J(\hat{p}_{i\mid k-1,opt})=0$ for some $i=1,\ldots,N$, where
$i_{0}(\hat{p}_{k-1,opt})$ is the index of the first occurrence that satisfies  $J(\hat{p}_{i_{0}\mid k-1,opt})=0$ in
$\hat{\boldsymbol{p}}_{k-1,opt}$;
\item $J(\hat{p}_{N\mid k})<\infty$ if $J(\hat{p}_{k-1}^{\ast})=0$ .
\end{enumerate} 

Applying $\hat{p}_{k}^{\ast}=\hat{p}_{1|k,opt}$ at each time $k$, the
optimal path $\hat{\boldsymbol{p}}^{\ast}=\hat{p}_{0}^{\ast}\hat{p}_{1}^{\ast}\ldots$
is guaranteed to satisfy the acceptance condition.
\label{thm:RHC_thm}
\end{thm}

\begin{IEEEproof}
% Following similar analysis in \cite{Cai2020c}, the energy function-based constraints ensure that $\hat{\boldsymbol{p}}^{\ast}=\hat{p}_{0}^{\ast}\hat{p}_{1}^{\ast}\ldots$
% intersects the accepting states $\mathcal{F}_{\mathcal{\hat{P}}}$
% infinitely which guarantees the satisfaction of the acceptance condition
% of $\hat{\mathcal{P}}$. Readers are referred to \cite{Cai2020c} for detailed
% control synthesis.
Consider a state $\hat{p}_{k-1}^{\ast}\in P,\forall k=1,2,\dots$ and 
$\textrm{Path}(\hat{p}_{k-1}^{\ast},N)$ represents the set of all possible paths starting from $\hat{p}_{k-1}^{\ast}$ with horizon $N$. Since not all predicted trajectories maximizing the utility function $\boldsymbol{U}(\boldsymbol{\hat{p}}_{k}),\boldsymbol{\hat{p}}_{k}\in\textrm{Path}(\hat{p}_{k-1}^{\ast},N)$ in (\ref{eq:P_kopt}) are guaranteed to satisfy the acceptance condition of $\mathcal{\hat{P}}$, additional constraints need to be imposed. The key idea about the design of the constraint for (\ref{eq:P_kopt}) is to ensure the energy of the states along the trajectory eventually decrease to zero. Therefore, we consider the following three cases. 
\begin{enumerate}
    \item Case 1: if $J(\hat{p}_{k-1}^{\ast})>0$ and $J(\hat{p}_{i\mid k-1,opt})\neq0$ for all $i=1,\dots,N$, the constraint $J(\hat{p}_{N\mid k})<J(\hat{p}_{N\mid k-1,opt})$ is enforced. The energy $J(\hat{p}_{k-1}^{\ast})>0$ indicates there exists a trajectory from $\hat{p}_{k-1}^{\ast}$ to $\mathcal{F}^{\ast}$, and 
    $J(\hat{p}_{i\mid k-1,opt})\neq0$ for all $i=1,\dots,N$ indicates $\boldsymbol{\hat{p}}_{k-1,opt}$ does not intersect $\mathcal{F}^{\ast}$. The constraint $J(\hat{p}_{N\mid k})<J(\hat{p}_{N\mid k-1,opt})$ enforces that the
    optimal predicted trajectory $\hat{p}_{N\mid k}$ must end at a state with lower energy than that of the previous predicted trajectory $\boldsymbol{\hat{p}}_{k-1,opt}$, which indicates the energy along $\boldsymbol{\hat{p}}_{k,opt}$ decreases at each iteration $k$.
    \item Case 2: if $J(\hat{p}_{i\mid k-1,opt})=0$ for some $i=1,\dots,N$, $\boldsymbol{\hat{p}}_{k-1,opt}$ intersects $\mathcal{F}^{\ast}$. Let $i_{0}(\hat{p}_{k-1,opt})$ be the index of the first occurrence in $\boldsymbol{\hat{p}}_{k-1,opt}$ where $J(\hat{p}_{i_{0}\mid k-1})=0$. The constraint $J(\hat{p}_{i_{0}(\hat{p}_{k-1,opt})-1\mid k})=0$ enforces the predicted trajectory at the current time $k$ to have energy 0 if the previous predicted trajectory contains such a state.
    \item Case 3: if $J(\hat{p}_{k-1}^{\ast})=0$, it indicates $\hat{p}_{k-1}^{\ast}\in\mathcal{F}^{\ast}$. The constraint $J(\hat{p}_{N\mid k})<\infty$ only requires the predicted trajectory $\boldsymbol{\hat{p}}_{k}$ ending at a state with bounded energy, where Cases 1 and 2 can then be applied to enforce the following sequence $\hat{p}_{k+1}^{\ast}\hat{p}_{k+2}^{\ast}\dots$ converging to $\mathcal{F}^{\ast}$. 
\end{enumerate}
\end{IEEEproof}

Since the environment is dynamic and unknown, the agent will update the environment according to the detected information at each time step. 
% If the predictive horizon $N$ is less than or equal to the sensor range $N_s$, then the local environment can be considered static. 
In addition, by selecting the predictive horizon $N$ to be less than or equal to the sensor range $N_s$,  we can ensure the existence of the solutions, since the local environment can be regarded as static.
As a result, lemmas in \cite{Ding2014} can be applied directly and the proof of the existence is omitted here.

Similar as \cite{Cai2020c}, the energy function based constraints (\ref{eq:P_kopt}) in Theorem \ref{thm:RHC_thm} ensure an optimal trajectory $\hat{\boldsymbol{p}}^{\ast}=\hat{p}_{0}^{\ast}\hat{p}_{1}^{\ast}\ldots$ is obtained which satisfies the acceptance condition.
Since the hard constraint is not relaxed, we can restrict the agent to avoid collisions at each time-step based on the sensor information. We assume the local information of WTS can be accurately updated such that the hard constraint is guaranteed. The system will return no solution in cases where no feasible trajectories satisfy the hard constraint, e.g., obstacles surrounding the agent.
Note that the optimality mentioned in this paper refers to local optimum since RHC controllers only optimize the objective within finite predictive steps.

The control synthesis of the MITL online motion planning strategy
is presented in the form of Algorithm \ref{alg:Controlsynthesis}.
Lines 2-3 are responsible for the offline initialization to obtain
an initial $\boldsymbol{J}$. The rest of Algorithm \ref{alg:Controlsynthesis}
(lines 4-16) is the online receding horizon control part 
executed at each time step. In Lines 4-6 the receding horizon control
is applied to determine $\hat{p}_{0}^{\ast}$ at time $k=0$. Since
the environment is dynamic and unknown, Algorithm \ref{alg:Controlsynthesis}
is applied at each time $k>0$ to update $\boldsymbol{J}$
based on local sensing in Lines 7-9. The RHC is then employed based
on the previously determined $\hat{p}_{k-1}^{\ast}$ to generate $\hat{\boldsymbol{p}}_{k,opt}$,
where the next state is determined as $\hat{p}_{k}^{\ast}=\hat{p}_{1|k,opt}$
in Lines 10-12. The transition from $\hat{p}_{k-1}^{\ast}$ to $\hat{p}_{k}^{\ast}$
applied on $\mathcal{\hat{P}}$ corresponds to the movement of the
agent at time $k$ from $\gamma_{\mathcal{T}}(\hat{p}_{k-1}^{\ast})$
to $\gamma_{\mathcal{T}}(\hat{p}_{k}^{\ast})$ on
$\mathcal{T}$ in Line 11. By repeating the process in lines 7-13, an optimal path $\hat{\boldsymbol{p}}^{\ast}=\hat{p}_{0}^{\ast}\hat{p}_{1}^{\ast}\ldots$
can be obtained that satisfies the acceptance condition of  $\mathcal{\hat{P}}$. 

\begin{algorithm}
\caption{\label{alg:Controlsynthesis}Control synthesis of MITL online motion
planning}

\small

\singlespacing

\begin{algorithmic}[1]

\Procedure {Input: } {The WTS $\mathscr{\mathcal{T=\textrm{\ensuremath{\left(Q,q_{0},\delta,\mathcal{AP},L,\mathcal{\omega}\right)}}}}$and
the relax TBA $\mathcal{\hat{A}}=(\hat{S},\hat{S}_{0},\mathcal{AP},\mathcal{L},X,I_{X},v_{c},v_{d},\hat{E},\hat{F})$
corresponding to the MITL formula $\phi=\phi_{h}\land\phi_{s}$}

{Output: } { the path $\hat{\boldsymbol{p}}^{\ast}=\hat{p}_{0}^{\ast}\hat{p}_{1}^{\ast}\ldots$
}

{Off-line Execution:}

\State Construct the relaxed product automaton $\hat{\mathcal{P}}=\mathcal{T\times\hat{A}}$

\State Construct $\mathcal{F}^{\ast}$ and initialize $\mathit{\boldsymbol{J}}$

{On-line Execution:}

\If { $\exists\hat{p_{0}}\in\hat{P_{0}}$ $J(\hat{p_{0}})<\infty$}

\State Solve for $\hat{\boldsymbol{p}}_{0,opt}$

\State $\hat{p}_{0}^{\ast}=\hat{p}_{1|0,opt}$ and $k\leftarrow1$

\While { $k>0$ }

\State Apply automaton update at $\hat{p}_{k-1}^{\ast}$ in Algorithm
\ref{alg:Automaton-Update} based on local sensing

\State Locally observe rewards $\boldsymbol{R}(\gamma_{\mathcal{T}}(\hat{p}_{k-1}^{\ast}))$

\State Solve for $\hat{\boldsymbol{p}}_{k,opt}$

\State Implement corresponding transitions on $\mathcal{\hat{P}}$
and $\mathcal{T}$

\State $\hat{p}_{k}^{\ast}=\hat{p}_{1|k,opt}$ and $k++$

\EndWhile

\Else{ There does not exist an accepting run from initial states}

\EndIf

\EndProcedure

\end{algorithmic}
\end{algorithm}

\textbf{Complexity Analysis}: Since the off-line
execution involves the computation of $\hat{\mathcal{P}}$, $\mathcal{F}^{\ast}$ and
the initial $\boldsymbol{J}$, its complexity is $O(\left|\mathcal{F}_{\hat{P}}\right|^{3}+\left|\mathcal{F}_{\hat{P}}\right|^{2}+\left|\hat{P}\right|^{2}\times\left|\mathcal{F}_{\hat{P}}\right|)$.
For online execution, since $\mathcal{F}^{\ast}$ remains the same
from Lemma 1, Algorithm 2 requires $\left|\hat{P} \right|$
runs of Dijkstra\textquoteright s algorithm. Suppose the number of
$\textrm{Sense}(\hat{p})$ is bounded by $\left|N_{1}\right|$, therefore,
the complexity of Algorithm 2 is at most $O(\left|N_{1}\right|\times\left|\hat{P}\right|+\left|\hat{P}\right|)$.
Suppose the number
of total transitions between states is $\left|\Delta_{\delta}\right|$.
In Algorithm 3, the complexity of recursive computation at each time
step is highly dependent on the horizon $N$ and is bounded by $\left|\Delta_{\delta}\right|^{N}$.
Overall, the maximum complexity of the online portion of RHC is $O(\left|N_{1}\right|\times\left|\hat{P}\right|+\left|\hat{P}\right|+\left|\Delta_{\delta}\right|^{N})$. 

\section{Case Studies\label{sec:Case}}

The simulation was implemented in MATLAB on a PC with 3.1 GHz Quad-core
CPU and 16 GB RAM. 
We demonstrate our framework using the Pac-Man setup
shown in Section \ref{sec:PF}. Consider an MITL specification $\phi=\phi_{h}\land\phi_{s}$,
where $\phi_{h}=\Square\lnot\mathtt{obstacle}$ and $\phi_{s}=\Square(\lnot\mathtt{grass})\land\Square\lozenge_{t<10}\mathtt{cherry}\land\Square(\mathtt{cherry}\rightarrow\lozenge_{t<20}\mathtt{pear)}.$
In English, $\phi_{h}$ means the agent has always to avoid obstacles,
and $\phi_{s}$ indicates the agent needs to repeatedly and sequentially
eat pears and cherries within the specified time intervals while avoiding
the grass. The tool \cite{Brihaye2017} allows converting MITL into TBA. Fig. \ref{fig:Snapshots} shows the snapshots during mission
operation. The simulation video is provided \footnote{\url{https://youtu.be/S_jfavmFIMo}}. 

\textbf{Simulation Results:} As for the priorities of violations, we set up that avoiding grass is more critical than eating
fruits within the specified time, i.e., we prefer to avoid discrete
violation rather than the continuous violation when $\phi_{s}$ is
infeasible. Therefore, we set the parameters $\alpha=0.8$ and $\beta=10$. The Pac-Man starts at the bottom
left corner and can move up, down, left, and right. In the maze, the
time-varying reward $R_{k}(q)$ is randomly generated at region
$q$ from a uniform distribution at time $k$. 

Since the WTS $\mathcal{T}$ has $\left|Q\right|=100$
states and the relaxed TBA $\mathcal{\hat{A}}$ has $\left|\hat{S}\right|=15$
states, the relaxed product automaton $\mathcal{\hat{P}}$ has $\left|\hat{P}\right|=1500$
states. The computation of $\mathcal{\hat{P}}$, the largest self-reachable
set $\mathcal{F}^{\ast}$, and the energy function took $0.62$s. The control
algorithm outlined in Algorithm \ref{alg:Controlsynthesis} is implemented
for $50$ time steps with horizon $N=4$. 

Fig. \ref{fig:Snapshots} shows the snapshots during mission operation.
Fig. \ref{fig:Snapshots} (a) shows that Pac-Man plans to reach cherry
within the specified time interval. Fig. \ref{fig:Snapshots} (b)
shows that $\phi_{s}$ is relaxed, and Pac-Man has two choices: go
straight to the left, pass the grass, and eat the pear  within the
specified time or go up first and then to the left to avoid the grass
and eat pear beyond the specified time. The former choice means discrete
violation while the latter means continuous violation. Since the avoidance of discrete violations has higher priority in  our algorithm, the agent chooses
the second plan as the predicted optimal path illustrated. Note that,
due to the consideration of dynamic obstacles, the deployment of black
blocks can vary with time. Fig. \ref{fig:Snapshots} (c) and (d) show
that on the second completion of the MITL task Pac-Man detects that
the cherry at the right top corner is blocked by obstacles and chooses
to eat the bottom one. 

\begin{figure}
\centering{}\includegraphics[scale=0.4]{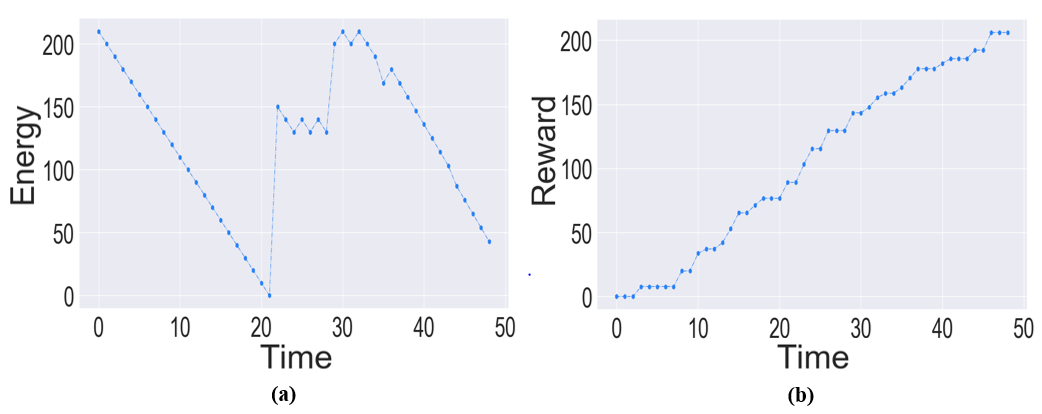}\caption{\label{fig:energy}The evolution of the energy function (a) and the
accumulative collected time-varying rewards\textcolor{black}{{} (b)}
during mission operation.}
\end{figure}

Fig. \ref{fig:energy} (a) shows the evolution
of the energy function during mission operation. Each time the energy
$J(\hat{p})=0$ in Fig. \ref{fig:energy} (a) indicates that an accepting
state has been reached, i.e., the desired task is accomplished for
one time. The jumps of energy from $t=30s$ to $35s$ (e.g., $t=30s$)
in Fig. \ref{fig:energy} (a) are due to the violation of the desired
task whenever the soft task is relaxed. Nevertheless, the developed
control strategy still guarantees the decrease of energy function
to satisfy the acceptance condition of $\mathcal{\hat{P}}$. Fig.
\ref{fig:energy} (b) shows the collected local time-varying rewards.

\textbf{Computation Analysis:} To demonstrate out algorithm's scalability and computational complexity, we repeat the control synthesis introduced above for
workspace with different sizes. The sizes of the resulted graph, WTS
$\mathcal{T}$, the relaxed product automaton $\mathcal{\hat{P}}$, and the meantime taken to solve the predicted trajectories at each
time-step are shown in Table I. We also analyze the effect
of horizon $N$ on the computation. From Table I, we can
see that in the cases with the same horizon $N$, the computational time increases gradually along with the increased workspace size. It is because trajectory updating involves
recomputing the energy function based on the updated environment knowledge. In this paper, the proposed RHC-based algorithm only
needs to consider the local optimization problem, and the energy
constraints will ensure global task satisfaction. Therefore, the
mean computation time at each time step does not increase significantly.
It shall be noted that in general RHC optimizations, the computations are influenced by the pre-defined
horizon $N$.

\begin{table}
\caption{The comparison of workspace size, horizon and computation time.}
\centering{}%
\begin{tabular}{|c|c|c|c|c|}
\hline 
$\begin{array}{c}
\textrm{Workspace}\\
\textrm{size}
\end{array}$ & $\begin{array}{c}
\textrm{Horizon}\\
N
\end{array}$ & $\begin{array}{c}
\mathcal{T}\\
\left|Q\right|
\end{array}$ & $\begin{array}{c}
\hat{\mathcal{P}}\\
\left|\hat{P}\right|
\end{array}$ & $\begin{array}{c}
\mathscr{\textrm{Mean}}\\
\textrm{time}(s)
\end{array}$\tabularnewline
\hline 
$10\times10$ & 4 & 100 & 1500 & 0.98\tabularnewline
\hline 
$10\times10$ & 6 & 100 & 1500 & 1.01\tabularnewline
\hline 
$10\times10$ & 8 & 100 & 1500 & 1.05\tabularnewline
\hline 
$30\times30$ & 4 & 900 & 13500 & 1.36\tabularnewline
\hline 
$30\times30$ & 6 & 900 & 13500 & 1.39\tabularnewline
\hline 
$30\times30$ & 8 & 900 & 13500 & 1.54\tabularnewline
\hline 
$50\times50$ & 4 & 2500 & 37500 & 2.91\tabularnewline
\hline 
$50\times50$ & 6 & 2500 & 37500 & 3.02\tabularnewline
\hline 
$50\times50$ & 8 & 2500 & 37500 & 3.60\tabularnewline
\hline 
\end{tabular}
\end{table} 

\section{Conclusion }

In this paper, we propose a control synthesis under hard and soft constraints given as MITL specifications. A relaxed timed product automaton is constructed for task relaxation consisting of task and time violations. An online motion planning strategy is synthesized with a receding horizon controller to deal with the dynamic and unknown environment and achieve multi-objective tasks. Simulation results validate the proposed approach.
% Since motion planning in
% an uncertain environment can be better modeled by Markov decision
% processes, future research will consider combining learning based
% methods and extending the current work to stochastic systems. 
% Since motion planning in
% an uncertain environment can be better formulated via Markov decision
% processes (MDP), future research will consider online robust planning for stochastic systems.
Future research will consider building the deterministic system online based on the real-time sensing information and develop online robust planning methods for stochastic systems.

\bibliographystyle{IEEEtran}
\bibliography{BibMaster}

\end{document}